\documentclass[11pt]{article}

\usepackage[preprint]{acl}

\usepackage{times}
\usepackage{latexsym}

\usepackage[T1]{fontenc}

\usepackage[utf8]{inputenc}

\usepackage{microtype}

\usepackage{inconsolata}

\usepackage{graphicx}
\usepackage{amsmath}
\usepackage{amssymb}
\usepackage{booktabs}
\usepackage{multirow}
\usepackage{enumitem}
\usepackage{subcaption}
\usepackage[most]{tcolorbox}

\DeclareMathOperator{\TopK}{TopK}

\newcounter{algobox}

\title{Self-Evolving World Models for LLM Agent Planning}

\author{Xuan Zhang$^1$ ~~~ Wenxuan Zhang$^2$ ~~~ See-Kiong Ng$^1$ ~~~ Yang Deng$^{3,*}$ \\
  $^1$National University of Singapore \\
  $^2$Singapore University of Technology and Design ~~~
  $^3$Singapore Management University \\ 
  \texttt{xuanzhang@u.nus.edu} \\
}

\begin{document}
\maketitle
\begingroup
\renewcommand{\thefootnote}{\fnsymbol{footnote}}
\footnotetext[1]{Corresponding author.}
\endgroup

\begin{abstract}
World models offer a principled way to equip long-horizon LLM agents with \emph{foresight}: predictions of action consequences before execution. However, unreliable foresight can be ignored, misused, or even degrade downstream decision-making. In this paper, we introduce \textsc{WorldEvolver}, a self-evolving world model framework that revises its deployment-time context while keeping the downstream agent and all model parameters frozen. \textsc{WorldEvolver} integrates three modules: (i) Episodic Memory, which exploits real action transitions through retrieval-based simulation; (ii) Semantic Memory, which extracts persistent heuristic rules from prediction-observation mismatches; and (iii) Selective Foresight, which filters low-confidence predictions before integrating them into agent reasoning context. We evaluate \textsc{WorldEvolver} on ALFWorld and ScienceWorld, measuring world model prediction accuracy on Word2World and downstream agent success rate on AgentBoard. Extensive experiments show that \textsc{WorldEvolver} achieves the highest prediction accuracy across three backbones and leads other world model baselines on downstream agent success rate, demonstrating that test-time memory revision enhances both predictive fidelity and planning performance.
\end{abstract}

\section{Introduction}
\label{sec:introduction}


LLM agents are typically improved through memory: reusing verbal feedback, retrieved experiences, skill libraries, or persistent context across interactions~\citep{shinn2023reflexion,wang2023voyager,packer2023memgpt}. A complementary paradigm is emerging through world models~\citep{li2025comprehensive,ding2025understanding,maes2026leworldmodel}, where agents improve not only by recalling past interaction experience, but also by anticipating future outcomes under candidate actions, analogous to learned environment models in model-based reinforcement learning~\citep{ha2018worldmodels,hafner2025dreamerv3}. Recent LLM-agent work follows this intuition through next-state prediction for web navigation~\citep{chae2025wma}, one-step visual web lookahead~\citep{gu2025webdreamer}, explicit prediction before ReAct-style action~\citep{fu2025preact}, and task knowledge models for text-game planning~\citep{qiao2024wkm}. These works suggest that world-model \textit{foresight} can serve as a useful complement to memory-based adaptation, particularly for planning and decision making in long-horizon tasks.

\begin{figure}[t]
  \centering
  \includegraphics[width=0.95\columnwidth]{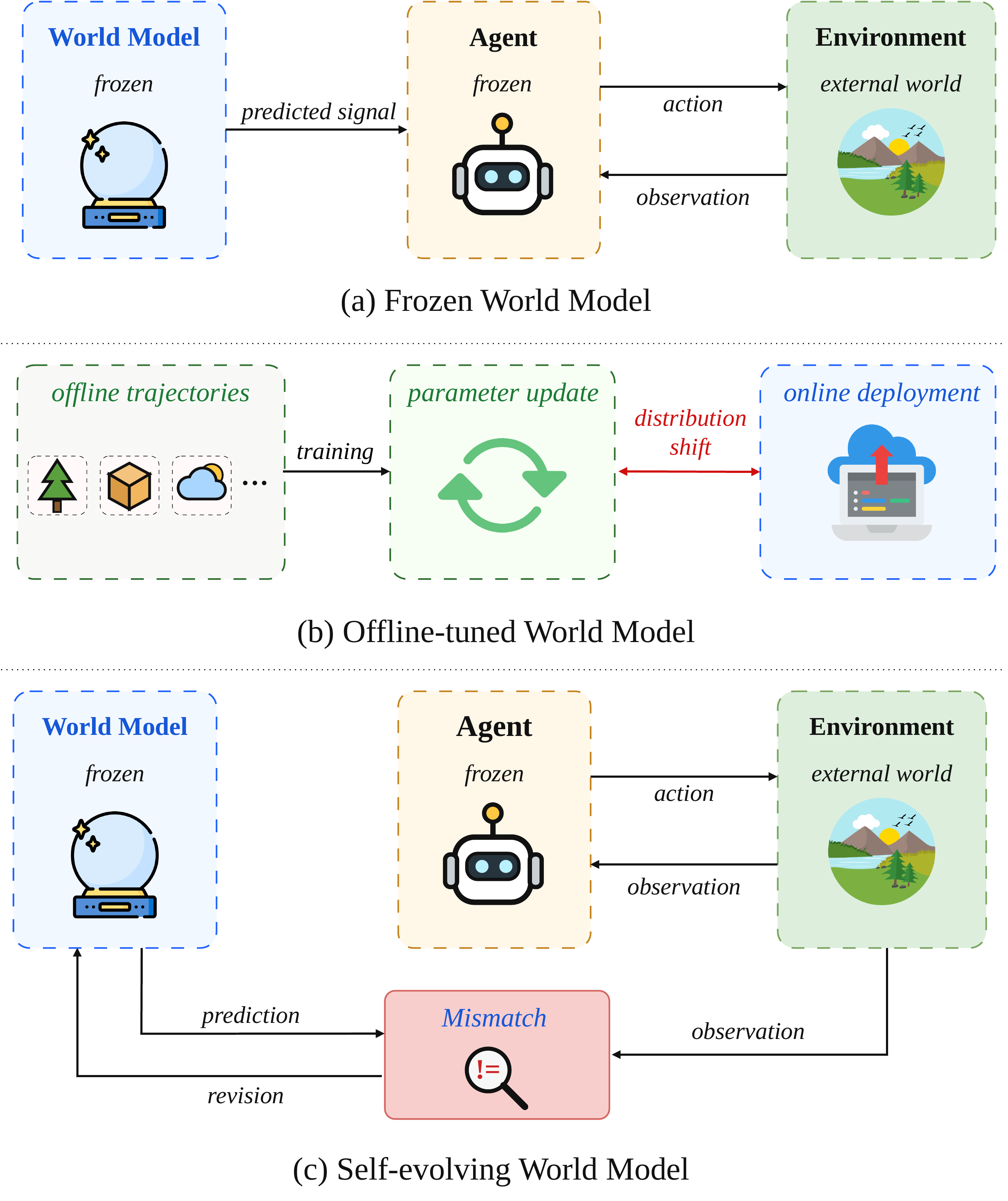}
  \caption{Contrast of different world models. Frozen (a) and offline-tuned (b) world models supply predictions to the agent without revising from deployment-time interaction; self-evolving (c) world models accumulate realized transitions and evolve through mismatches between predicted and observed outcomes.}
  \label{fig:wm_patterns}
  \vspace{-0.5cm}
\end{figure}

However, the reliability of \textit{foresight} is not static. Deployed agents continually face evolving environments and new task instances, creating distribution shifts analogous to the sim-to-real gap in robotics~\citep{tobin2017domainrandomization}.
As a result, a frozen world model (Figure~\ref{fig:wm_patterns}(a)) suffers from such distribution shifts and can mispredict future transitions. At the same time, absorbing each mismatch through gradient-based parameter updates (Figure~\ref{fig:wm_patterns}(b)) is a poor fit for online deployment: such updates incur high computation costs at LLM scale and can introduce side effects such as over-editing or catastrophic forgetting~\citep{zheng2023ike,yao2023editing,hartvigsen2023aging}. 

This makes self-evolution~\citep{qiu2026self,chu2026worldmodelsurvey} a fundamental requirement for deployed world models (Figure~\ref{fig:wm_patterns}(c)): they should detect mismatches between predicted and observed outcomes and adapt accordingly.
Meanwhile, the agent-environment loop already exposes reusable evidence: realized transitions record what actually happened, while prediction-observation mismatches indicate what the world model misunderstood. Retaining these signals as explicit context offers an auditable alternative to repeated parameter updates, so later predictions can condition on deployment-time evidence to generate more reliable \textit{foresight} without changing model weights.

\begin{figure}[t]
  \centering
  \includegraphics[width=0.98\columnwidth]{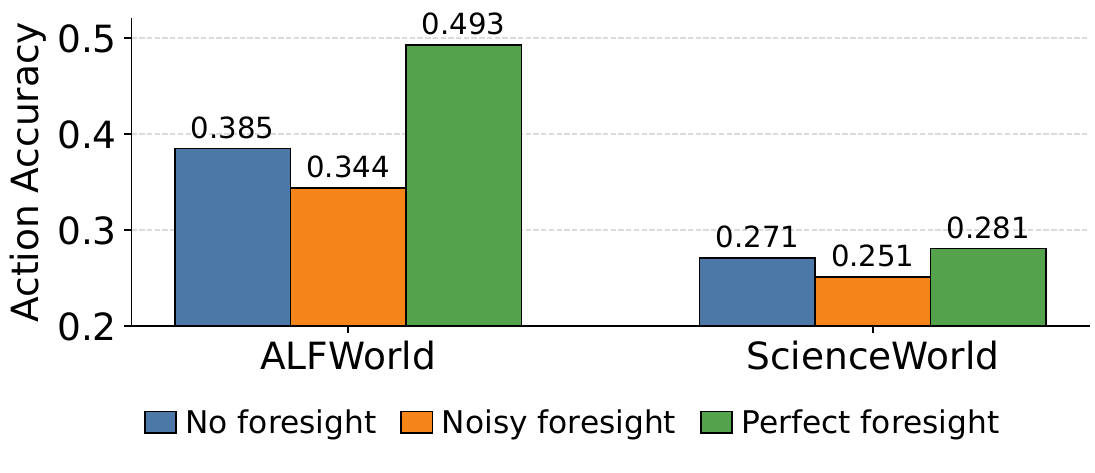}
  \caption{Preliminary oracle study with Gemma-4-26B-A4B on the Word2World evaluation set~\citep{li2025wordtoworld}. The ReAct agent receives no \textit{foresight}, noisy \textit{foresight}, or perfect \textit{foresight}, and generated actions are scored against teacher actions by exact action accuracy.}
  \label{fig:prelim_stage1}
  \vspace{-0.5cm}
\end{figure}

Even once such evidence is retained, \textit{foresight} remains an action-conditioning signal: once rendered to the agent, it can change the next action. Prior work shows that current agents can ignore, misuse, or even be harmed by world-model simulations~\citep{qian2026foresight}, echoing model-based RL evidence that learned rollouts should be trusted selectively under model error~\citep{janner2019mbpo}. Similarly, recent adaptive-lookahead work further suggests that useful imagination depends on when and how far the agent should simulate, rather than on fixed-horizon rollouts~\citep{liu2026imaginethenplan}.
The controlled oracle diagnostic in Figure~\ref{fig:prelim_stage1} provides supporting evidence under a fixed agent and backbone: noisy \textit{foresight} hurts action accuracy, while oracle \textit{foresight} improves it.

These observations motivate \textsc{WorldEvolver}: a standalone self-evolving world model that continuously revises its deployment-time context while the downstream agent and all model parameters remain frozen. It addresses three successive deployment-time challenges through a three-module design: grounding predictions in realized transitions when parametric knowledge is insufficient, abstracting individual prediction-observation mismatches into reusable knowledge, and filtering unreliable \textit{foresight} before it can condition the agent. This structure supports online adaptation.

Concretely, our proposed \textsc{WorldEvolver} couples three complementary mechanisms in an executed-action-aligned loop. \textbf{Episodic Memory} serves as the exploitation component that reuses accumulated action-transition experience through retrieval-based simulation, while \textbf{Semantic Memory} acts as the exploration component that turns prediction-observation mismatches into persistent heuristic knowledge. To mitigate the risk of unreliable \textit{foresight}, \textbf{Selective Foresight} filters low-confidence predictions before exposing them to the frozen agent. If \textit{foresight} changes the draft action, the world model predicts the revised action again and, after execution, stores the realized transition in $M_E$ and updates $M_S$ from the corresponding prediction-observation mismatch.
In summary, our contributions are as follows:
\begin{itemize}[leftmargin=*, itemsep=0.1em]
    \item
    We introduce \textsc{WorldEvolver}, a standalone self-evolving world model for LLM agents that revises the deployment-time world-model context while the agent and all model parameters remain frozen during environmental interaction.
    \item We instantiate this memory-centric \textit{foresight} framework as an executed-action-aligned loop with three mechanisms: Episodic Memory retrieves realized transitions, Semantic Memory accumulates mismatch-derived rules, and Selective Foresight filters unreliable predictions before they reach the agent.
    \item
    We benchmark the \textsc{WorldEvolver} framework against RAWM-$\phi$ and ITP-I on Word2World~\citep{li2025wordtoworld}, ALFWorld~\citep{shridhar2021alfworld}, and ScienceWorld~\citep{wang2022scienceworld}, evaluating both world-model prediction alignment with future observations and downstream planning improvements from the generated \textit{foresight}.
\end{itemize}

\section{Related Work}
\label{sec:related_work}

\textbf{World Models For LLM Agents.}~
World-model \textit{foresight} extends the model-based reinforcement learning lineage~\citep{ha2018worldmodels,hafner2025dreamerv3} to language agents. Existing systems instantiate this idea by using an LLM as both planner and simulator~\citep{hao2023rap}, training next-state predictors for web navigation~\citep{chae2025wma}, adding explicit prediction before action~\citep{fu2025preact}, or learning task-level world knowledge for text-game planning~\citep{qiao2024wkm}.
Other work improves \textit{foresight} through offline training or joint optimization, such as co-training agents and world models~\citep{fang2025webevolver}, retrieval-augmented world model learning~\citep{yang2025rawm}, and synthetic-environment training~\citep{ding2026dynaweb}. RAWM-$\phi$ retrieves from a fixed transition library with a PPO-trained head~\citep{yang2025rawm}. AutoManual constructs an online manual for the planner~\citep{chen2024automanual}, while WALL-E 2.0 builds symbolic transition knowledge for model-predictive control~\citep{zhou2025walle2}. Imagine-then-Plan adapts the imagination horizon~\citep{liu2026imaginethenplan}. A complementary lesson from episodic-control and language-agent memory systems is that accumulated interaction experience can ground later decisions through retrieved histories~\citep{blundell2016model,pritzel2017nec,deng2024multi,zheng2024synapse,zhong2024memorybank,zhou2024trad,liu2025cer}. \textsc{WorldEvolver} starts with empty memory, accumulates executed transitions and mismatch-derived textual rules online without parameter updates, confidence-gates one-step predictions, aligns memory updates with the executed action, and leaves the downstream policy frozen.

\noindent\textbf{Self-Evolution.}~
Recent work increasingly studies self-evolving agents, where interaction improves behavior through verbal feedback, skill libraries, distilled experience, or persistent context~\citep{wang2023voyager,packer2023memgpt,zhao2024expel}.
A growing line of \emph{fully} autonomous approaches removes human supervision and bootstraps agents from zero or minimal data. Challenger-solver methods co-evolve a task proposer and a solver, with the proposer generating increasingly difficult tasks~\citep{huang2025rzero,yu2025guided,xia2025agent0,yue2026drzero}. WebRL instead adapts an online curriculum from the web agent's current performance~\citep{qi2025webrl}. Early Experience and DreamGym improve agents by collecting or synthesizing interaction data~\citep{zhang2025early,chen2025dreamgym}, while Co-Evolving Agents and CURE use failures or mutual evaluation to improve specialized learners~\citep{jung2025coevolving,wang2025cure}. Several works also couple agents with co-evolving task generators or environment simulators that adapt to the agent's frontier~\citep{guo2025genenv}. Closer to our setting, a few recent efforts begin to evolve a learned
world model alongside the agent, either by retraining it on environment
rollouts via self-supervised RL~\citep{yu2026rwml,ding2026dynaweb},
dynamically updating an abstracted state model during
exploration~\citep{kim2025coex}, or alternating updates between neural and
symbolic components~\citep{zhao2026nesys}. However, most existing methods evolve the agent policy or external context, rather than the world model that supports future prediction. As a result, they do not directly address how predictive models should adapt under changing environments or unreliable \textit{foresight}.

\section{Methodology}
\label{sec:methodology}

\subsection{Problem Formulation}
\label{sec:method:problem}

We formulate each task as a partially observed interaction process $(\mathcal{S},\mathcal{A},\mathcal{O},\mathcal{T})$, where $\mathcal{S}$ is the environment state space, $\mathcal{A}$ is the action space, $\mathcal{O}$ is the observation space, and $\mathcal{T}:\mathcal{S}\times\mathcal{A}\rightarrow\mathcal{S}$ is the transition function. At time step $t$, the agent 
cannot directly access the hidden environment state. Instead, it observes a textual interaction state:
$$s_t=(o_1,a_1,\ldots,o_{t-1},a_{t-1},o_t),$$ 
where $o_i \in \mathcal{O}$ and $a_i \in \mathcal{A}$ denote observations and actions, respectively. Given the current state $s_t$, the agent policy generates an action:
$$a_t\sim\pi_{\theta}(\cdot\mid s_t).$$ 
A world model predicts future $K$-step observations from the current state and a candidate action: 
$$(\hat{o}_{t+1},\ldots,\hat{o}_{t+K}) \sim W_{\theta}(\cdot\mid s_t,a_t).$$ 
We mainly focus on one-step \textit{foresight} because 
the next predicted observation $\hat{o}_{t+1}$ from the world model directly influences the current action decision of the agent, while the realized observation $o_{t+1}$ immediately provides supervision on whether the prediction was reliable.
The goal is therefore to 
select and improve $\hat{o}_{t+1}$ through deployment-time continual evolution, while keeping both the agent policy $\pi_{\theta}$ and world model $W_{\theta}$ frozen.

\subsection{\textsc{WorldEvolver}}
\label{sec:method:arch}

\begin{figure*}[t]
  \centering
  \includegraphics[width=0.98\textwidth]{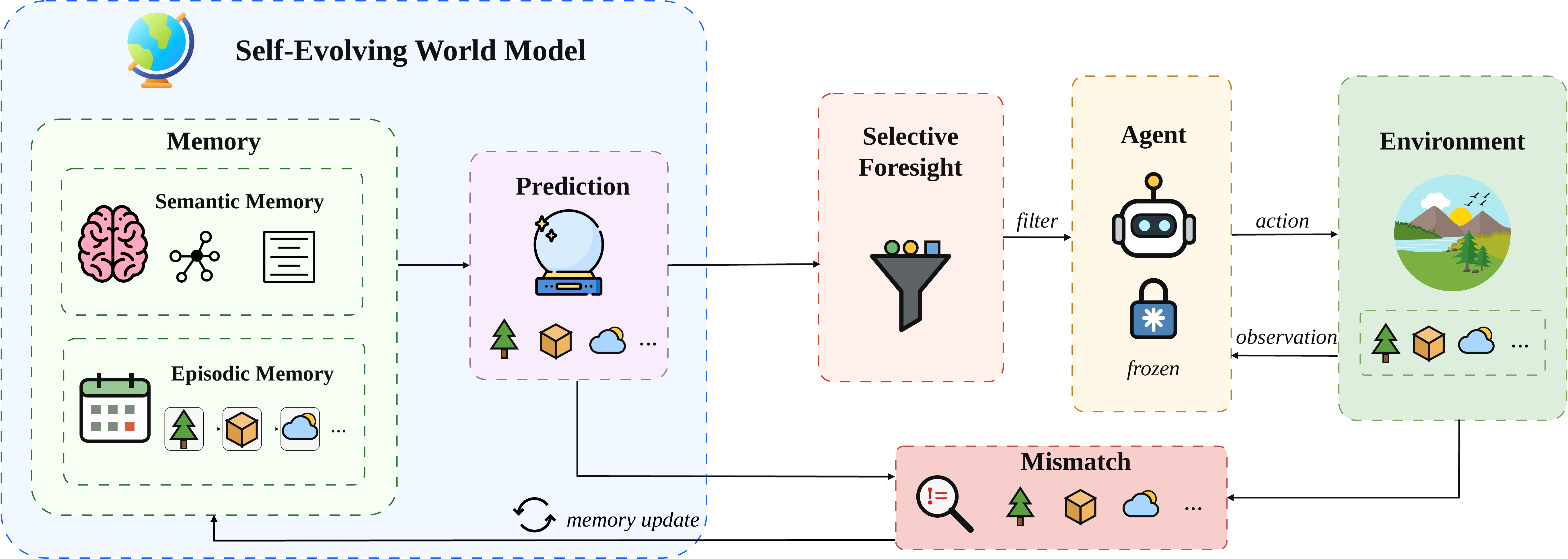}
  \caption{Overview of \textsc{WorldEvolver}. A frozen world model produces action-conditioned predictions using Episodic Memory for exploitation through retrieval-based simulation over previous action transitions and Semantic Memory for exploration through persistent heuristic-rule discovery from prediction-observation mismatches. Selective Foresight filters the prediction before it conditions the frozen agent. 
  }
  \label{fig:pipeline}
  \vspace{-0.3cm}
\end{figure*}


\textsc{WorldEvolver} addresses a central challenge in world-model-based agents: predicted futures can improve decision making, but unreliable \textit{foresight} may also mislead the agent. As illustrated in Figure \ref{fig:pipeline}, rather than updating model parameters, \textsc{WorldEvolver} evolves the evidence provided to the frozen world model at inference time.

At step $t$, the frozen world model $W_{\theta}$ is augmented with a non-parametric memory store:
$$
M_t = (M_E^t, M_S^t),
$$
where $M_E^t$ denotes Episodic Memory and $M_S^t$ denotes Semantic Memory. The world model conditions on the current task context, observation, candidate action, and retrieved memory to generate predictions that are gated by confidence.

Inspired by the classical distinction between episodic and semantic memory~\citep{tulving1972episodic}, Episodic Memory stores concrete interaction experiences, while Semantic Memory stores abstract, reusable knowledge. In \textsc{WorldEvolver}, Episodic Memory supports exploitation by recalling relevant transitions, whereas Semantic Memory supports exploration and uses prediction-observation mismatches as its update signal.


\paragraph{Episodic Memory}
Episodic Memory stores concrete interaction experiences. The key intuition is that previous transitions can provide useful grounding for predicting what may happen after a similar action in the current environment state. Prior work on language-agent memory shows that retrieved trajectories and replayed experiences can improve decision making by grounding new actions in previous interactions rather than relying only on abstract instructions.
The episodic memory contains realized transitions $M_E^t=\{(o_i,a_i,o_{i+1})\}_{i<t}$. Given a candidate action $a_t$ and retrieval size $k_{M_E}$, it 
retrieves the $k_{M_E}$ most similar past transitions:
$$M_{E,k_{M_E}}^t(a_t)=\TopK^{k_{M_E}}_{(o_i,a_i,o_{i+1})\in M_E^t}\operatorname{sim}(a_t,a_i).$$
It then renders each selected item as raw text containing the previous observation, action, and next observation in the context. The similarity function $\operatorname{sim}$ is defined as the Jaccard score over the open-vocabulary action token set. Since new memory records are appended only after execution, retrieval at step $t$ only relies on previously accumulated experience, with the episodic memory updated as
$$M_{E}^{t+1} = M_E^t \cup \{(o_t, a_t, o_{t+1})\}.$$

\paragraph{Semantic Memory}
The semantic memory converts prediction-observation mismatches into persistent textual heuristics without updating model parameters. Instead of treating mismatches as failures of the world model weights, we interpret them as feedback on the contextual memory.
Such mismatches provide correction evidence for improving future simulations. We store these corrections as $M_S^t=\{(r_i,e_i)\}_{i=1}^{|M_S^t|}$, where each $r_i$ is a heuristic rule with evidence score $e_i\in\mathbb{R}$.

Before applying the textual revision, we first compare predictions and observations in a factorized state space. The key comparison is therefore not whether two observations share the same surface wording, but whether they describe the same objects, relations, and actions. For example, the observation \textit{``The fridge 1 is open. In the fridge 1, you see an apple 1.''} can be factorized into tuples such as \textit{(`fridge 1', `is', `open')} and \textit{(`apple 1', `in', `fridge 1')}. Following \citet{hao2023rap} and~\citet{shen2026statefactory}, we use a mapping function $g$ to transform raw observation text into factorized tuples, producing $\hat{z}_{t+1}=g(\hat{o}_{t+1})$ and $z_{t+1}=g(o_{t+1})$.
The revision process therefore follows the pipeline
\begin{equation*}
\begin{split}
    &(s_t,a_t)\xrightarrow{W_\theta}\hat{o}_{t+1}, \\
    &(\hat{o}_{t+1},o_{t+1})\xrightarrow{g}(\hat{z}_{t+1},z_{t+1})\xrightarrow{\text{LLM critic}}r_i,
\end{split}
\end{equation*}
where the final stage produces textual feedback on the contextual memory rather than updating the world model parameters. When $\hat{z}_{t+1}\neq z_{t+1}$, the mismatch is treated as a failure case, and the LLM critic transforms it into candidate textual rules $r_i$. Each rule is associated with an evidence score $e_i$, initialized to $1$, which is updated by $\pm 1/|M_S|$ depending on whether the rule is supported or contradicted by the factorized-tuple comparison on subsequent observations. Only rules with $e_i>0$ are added to the context. The resulting rule-evidence pairs are collected as $\Delta M_S^t$, and the semantic memory is updated incrementally as
$$
M_S^{t+1}=M_S^t\cup\Delta M_S^t.
$$
Following batch semantic-gradient updates for language-based agent systems~\citep{wang2024correctly}, Semantic Memory can accumulate a mini-batch of $k_{M_S}$ mismatch cases before revising the rendered rule set. In this variant, the LLM critic produces $\Delta M_S^t$ as the aggregated rule-evidence updates over the mini-batch of mismatches.
Thus, Semantic Memory is the exploration branch: it turns failures into inspectable prompt-level knowledge without gradient updates to $W_\theta$.

\paragraph{Selective Foresight}

Although memory can improve prediction quality, unreliable \textit{foresight} may still mislead the downstream agent. As shown in Figure~\ref{fig:prelim_stage1}, noisy predictions can degrade decision making more than providing no \textit{foresight} at all~\citep{janner2019mbpo,qian2026foresight}. This raises a practical question: should the agent always trust the predicted future, or should unreliable predictions be filtered before they influence action selection?

Selective Foresight addresses this problem by exposing only sufficiently confident predictions to the agent policy. Suppose the world model generates a predicted observation sequence tokenized as $y_{1:n}$. 
When token probabilities are available from the backend model, we first compute the average token-level log probability:
$$\ell_t=\frac{1}{n}\sum\nolimits_{i=1}^{n} \log p_{\theta}(y_i\mid y_{<i},s_t,a_t,M_t),$$
and convert it into a normalized confidence score: $q_t=\exp(\ell_t)\in(0,1]$. 
This score corresponds to the geometric mean token probability of the output. The final agent-visible \textit{foresight} is defined as
\begin{equation*}
F_t =
\begin{cases}
\hat{o}_{t+1},
&
q_t \ge \tau,
\\
\varnothing,
&
q_t < \tau,
\end{cases}
\end{equation*}
where $\tau$ denotes the confidence threshold.

Selective Foresight therefore acts as an abstention mechanism based on prediction confidence, reducing the risk that unreliable simulations negatively influence downstream decision making.

\subsection{Agent Planning with World Models}
\label{sec:method:planning}

\begin{figure}[t]
  \centering
  \refstepcounter{algobox}\label{alg:worldevolver}
  \fbox{\begin{minipage}{0.94\columnwidth}
  \small
  \textbf{Algorithm~\thealgobox\ \textsc{WorldEvolver} Update}\\
  \noindent\rule{\linewidth}{0.35pt}\\
  \textbf{Input:} agent-visible state $s_t$, observation $o_t$, policy $\pi_{\theta}$, world model $W_{\theta}$,
  memories $M_E^t,M_S^t$, retrieval size $k_{M_E}$, semantic batch size $k_{M_S}$, threshold $\tau$.\\
  \textbf{Output:} executed action $a_t$ and updated memories $M_E^{t+1},M_S^{t+1}$.\\
  \vspace{0.2em}
  \begin{tabular}{@{}r p{0.83\linewidth}@{}}
  \multicolumn{2}{@{}l}{// Draft and predict}\\
  1: & Sample draft action $a_t^{(0)}\sim\pi_{\theta}(\cdot\mid s_t)$.\\
  2: & Retrieve $M_{E,k_{M_E}}^t(a_t^{(0)})$ from $M_E^t$ by action-token Jaccard score.\\
  3: & Query $W_{\theta}$ on $(s_t,a_t^{(0)},M_{E,k_{M_E}}^t(a_t^{(0)}),M_S^t)$ to obtain $(\hat{o}_{t+1},q_t)$.\\
  \multicolumn{2}{@{}l}{// Selective \textit{foresight}}\\
  4: & Set $F_t\leftarrow\hat{o}_{t+1}$ if $q_t\ge\tau$; otherwise $F_t\leftarrow\varnothing$.\\
  5: & Sample $a_t^{(1)}\sim\pi_{\theta}(\cdot\mid s_t,F_t)$.\\
  6: & Set executed action $a_t\leftarrow a_t^{(1)}$.\\
  \multicolumn{2}{@{}l}{// Align prediction with executed action}\\
  7: & If $a_t\ne a_t^{(0)}$, obtain a new $\hat{o}_{t+1}$ by querying $W_{\theta}$ on $(s_t,a_t,M_{E,k_{M_E}}^t(a_t),M_S^t)$. \\
  \multicolumn{2}{@{}l}{// Execute and update memory}\\
  8: & Execute $a_t$ in the environment and observe $o_{t+1}$.\\
  9: & Set $M_E^{t+1}\leftarrow M_E^t\cup\{(o_t,a_t,o_{t+1})\}$.\\
  10: & Compute or accumulate $\Delta M_S^t$ of size $k_{M_S}$ from $(\hat{o}_{t+1},o_{t+1})$ and update $M_S^{t+1}$.\\
  11: & Return $a_t,M_E^{t+1},M_S^{t+1}$.\\
  \end{tabular}
  \end{minipage}}
  \vspace{-0.5cm}
\end{figure}

Algorithm~\ref{alg:worldevolver} shows one closed-loop planning step. The agent (1) samples a draft action $a_t^{(0)}$ from the frozen policy, (2) retrieves $k_{M_E}$ episodic transitions for this action, and (3) asks the frozen world model to predict the consequence of $a_t^{(0)}$ under the current memory context. Selective Foresight then (4) converts the prediction into an agent-visible signal: the predicted observation is passed to the policy only when its confidence $q_t$ exceeds the threshold $\tau$; otherwise, no \textit{foresight} is provided to the policy. The agent policy subsequently (5-6) samples the executed action $a_t = a_t^{(1)}$ conditioned on $(s_t,F_t)$.

Because the final action may differ from the draft action used for the initial prediction, \textsc{WorldEvolver} aligns the learning signal with the action actually executed in the environment. When $a_t\ne a_t^{(0)}$, the world model (7) is queried once more with $a_t$ to obtain $\hat{o}_{t+1}$.
This second query ensures that Semantic Memory is updated from the mismatch between the prediction for the executed action and the realized observation. After (8) executing $a_t$, \textsc{WorldEvolver} (9) appends the realized transition $(o_t,a_t,o_{t+1})$ to Episodic Memory. It then compares the executed-action prediction $\hat{o}_{t+1}$ with the realized observation $o_{t+1}$; if they differ after factorization, the LLM critic (10) produces rule updates $\Delta M_S^t$. Finally, the algorithm (11) returns the action and updated non-parametric memories. This preserves action-level consistency.

\begin{table*}[t]
\centering
\footnotesize
\resizebox{\textwidth}{!}{%
\begin{tabular}{@{}lcccccc@{}}
\toprule
\multirow{2}{*}{Setting} & \multicolumn{3}{c}{ALFWorld} & \multicolumn{3}{c}{ScienceWorld} \\
\cmidrule(lr){2-4}\cmidrule(l){5-7}
& Exact Match & Token F1 & Cosine Similarity & Exact Match & Token F1 & Cosine Similarity \\
\midrule
\multicolumn{7}{@{}c@{}}{\textbf{Gemma-4-26B-A4B}} \\
\midrule
Zero-Shot & 3.60 & 35.48 & 67.91 & 0.41 & 16.42 & 52.45 \\
RAWM-$\phi$ \cite{yang2025rawm} & 20.06 & 48.13 & 71.30 & 14.93 & 27.31 & 56.50 \\
ITP-I \cite{liu2026imaginethenplan} & 1.46 & 32.48 & 66.10 & 0.39 & 11.50 & 47.69 \\
\textsc{WorldEvolver} (w/o $M_E$) & 7.53 & 38.08 & 69.06 & 2.71 & 19.29 & 55.04 \\
\textsc{WorldEvolver} (w/o $M_S$) & 47.16 & 72.61 & 78.88 & 34.65 & 46.93 & 66.51 \\
\textsc{WorldEvolver} & \textbf{52.88} & \textbf{76.75} & \textbf{80.13} & \textbf{51.55} & \textbf{62.43} & \textbf{73.85} \\
\midrule
\multicolumn{7}{@{}c@{}}{\textbf{Qwen3.5-9B}} \\
\midrule
Zero-Shot & 1.58 & 34.06 & 66.39 & 0.59 & 12.72 & 49.27 \\
RAWM-$\phi$ \cite{yang2025rawm} & 14.41 & 38.63 & 66.31 & 2.76 & 16.53 & 49.76 \\
ITP-I \cite{liu2026imaginethenplan} & 0.00 & 11.22 & 52.88 & 0.00 & 6.68 & 41.94 \\
\textsc{WorldEvolver} (w/o $M_E$) & 2.04 & 33.80 & 65.81 & 0.90 & 13.98 & 50.40 \\
\textsc{WorldEvolver} (w/o $M_S$) & 34.86 & 61.56 & 74.34 & 28.92 & 44.84 & 65.08 \\
\textsc{WorldEvolver} & \textbf{37.04} & \textbf{62.38} & \textbf{74.64} & \textbf{29.82} & \textbf{44.88} & \textbf{65.15} \\
\midrule
\multicolumn{7}{@{}c@{}}{\textbf{Gemma-4-31B}} \\
\midrule
Zero-Shot & 2.71 & 38.42 & 69.90 & 7.58 & 25.28 & 58.55 \\
RAWM-$\phi$ \cite{yang2025rawm} & 34.33 & 57.49 & 72.66 & 32.84 & 42.38 & 64.84 \\
ITP-I \cite{liu2026imaginethenplan} & 1.36 & 33.61 & 67.64 & 0.56 & 11.91 & 49.63 \\
\textsc{WorldEvolver} (w/o $M_E$) & 6.73 & 41.30 & 71.72 & 13.34 & 30.90 & 61.32 \\
\textsc{WorldEvolver} (w/o $M_S$) & 56.27 & 80.02 & 81.21 & 56.74 & 66.50 & 76.00 \\
\textsc{WorldEvolver} & \textbf{56.41} & \textbf{80.87} & \textbf{81.39} & \textbf{62.03} & \textbf{71.60} & \textbf{78.42} \\
\bottomrule
\end{tabular}
}
\caption{World model prediction accuracy on Word2World; higher is better for all metrics. \textit{w/o $M_E$} removes episodic memory, while \textit{w/o $M_S$} removes semantic memory. All memories $M_t$ are initialized empty and updated online during interaction. Unless otherwise specified, \textsc{WorldEvolver} here uses $k_{M_E}{=}5$ and $k_{M_S}{=}1$.}

\label{tab:stage2_main}
\vspace{-0.3cm}
\end{table*}

\section{Experiment}
\label{sec:experiment}
We evaluate \textsc{WorldEvolver} along two complementary axes. First, \textbf{World Model Prediction} (Section~\ref{sec:experiments:prediction}) measures how accurately the model predicts future observations relative to real environment transitions. Second, \textbf{Agent Planning} (Section~\ref{sec:experiments:agent_planning}) evaluates whether these world models improve closed-loop task performance for agents. Finally, Section~\ref{sec:discussion} discusses the effects of memory hyperparameters and online continual learning, with additional analyses provided in Appendix~\ref{sec:appendix:evaluation_analysis}.

\begin{table*}[t]
\centering
\footnotesize
\resizebox{\textwidth}{!}{%
\begin{tabular}{@{}llcccc@{}}
\toprule
\multirow{2}{*}{Agent} & \multirow{2}{*}{Setting} & \multicolumn{2}{c}{ALFWorld} & \multicolumn{2}{c}{ScienceWorld} \\
\cmidrule(lr){3-4}\cmidrule(l){5-6}
& & Gemma-4-26B-A4B & GPT-5.4-mini & Gemma-4-26B-A4B & GPT-5.4-mini \\
\midrule
\multirow{5}{*}{ReAct}
& \textit{w/o World Model} & \textit{23.88} & \textit{49.25} & \textit{44.44} & \underline{\textit{65.56}} \\
& RAWM-$\phi$ \cite{yang2025rawm} & 22.39 & 41.79 & 43.33 & 57.78 \\
& ITP-I \cite{liu2026imaginethenplan} & 25.37 & 38.81 & 34.44 & 60.00 \\
& \textsc{WorldEvolver} w/o $F_t$ & 24.63 & 43.28 & 46.67 & 62.22 \\
& \textsc{WorldEvolver} w/ $F_t$ & \underline{\textbf{26.12}} & \underline{\textbf{50.75}} & \underline{\textbf{52.22}} & \textbf{63.33} \\
\midrule
\multirow{5}{*}{ReflAct}
& \textit{w/o World Model} & \textit{26.12} & \underline{\textit{50.00}} & \textit{42.22} & \textit{60.00} \\
& RAWM-$\phi$ \cite{yang2025rawm} & 20.15 & 42.54 & 41.11 & 58.89 \\
& ITP-I \cite{liu2026imaginethenplan} & 23.13 & 30.60 & 37.78 & 58.89 \\
& \textsc{WorldEvolver} w/o $F_t$ & 24.63 & 44.78 & 48.89 & 62.22 \\
& \textsc{WorldEvolver} w/ $F_t$ & \underline{\textbf{27.61}} & \textbf{47.01} & \underline{\textbf{50.00}} & \underline{\textbf{63.33}} \\
\bottomrule
\end{tabular}
}
\caption{Agent planning success rate; higher is better. w/ and w/o $F_t$ denote with and without selective foresight. Underlines denote the best overall setting, and bold denotes the best setting among world-model-based methods.}
\vspace{-0.3cm}
\label{tab:stage3_sr}
\end{table*}

\subsection{Setups}
\label{sec:experiments:setups}
This subsection summarizes the experimental setup, and additional details are provided in Appendix~\ref{sec:appendix:experimental_setups}.

\paragraph{Datasets}
We conduct evaluations on both world model prediction and agent planning. To evaluate the alignment between prediction and ground truth, we adopt the Word2World Benchmark~\citep{li2025wordtoworld}, which provides transition datasets for ALFWorld~\citep{shridhar2021alfworld} and ScienceWorld~\citep{wang2022scienceworld}. The test split contains 195 trajectories for each environment. In agent planning, we use AgentBoard~\citep{ma2024agentboard}, with 134 ALFWorld tasks and 90 ScienceWorld tasks. Each configuration runs $L{=}5$ trials per task, with a maximum of 30 steps per trial.

\paragraph{Baselines}
We define each comparison by the foresight provided by the world model while keeping both the agent and backbone model fixed. We consider the following baselines: \textbf{Zero-Shot}, \textbf{RAWM-$\phi$}~\citep{yang2025rawm}, and \textbf{ITP-I}~\citep{liu2026imaginethenplan}.

\paragraph{Agents}
We apply two agent types with distinct planning styles to test whether the world model generalizes across reasoning paradigms: \textbf{ReAct}~\citep{yao2023react} and \textbf{ReflAct}~\citep{kim2025reflact}.

\paragraph{Evaluation Metrics}
Prediction metrics measure whether the world model matches the next observation; planning metrics measure whether the exposed signal helps the agent complete tasks.
\begin{itemize}[leftmargin=*, itemsep=0.1em]
    \item \textbf{World model prediction:} (1) Exact Match uses normalized string matching between predicted and reference observations. (2) Token F1 measures lexical overlap after tokenization, micro-averaged across all examples. (3) Cosine Similarity measures semantic similarity using \texttt{Qwen3-Embedding-8B}~\citep{zhang2025qwen3embedding} embeddings in the same retrieval space.
    \item \textbf{Agent planning:} 
    We report Success Rate, defined as whether the agent completes the task within the allowed interaction budget, and aggregate results using best-of-$L$ across trials.
\end{itemize}

\paragraph{Implementation Details}
\label{sec:experiments:implementation}
World model prediction uses \texttt{Qwen3.5-9B}~\citep{qwen2026qwen35}, \texttt{Gemma-4-26B-A4B}, and \texttt{Gemma-4-31B}~\citep{googledeepmind2026gemma4}. Agent planning evaluation uses \texttt{Gemma-4-26B-A4B} and \texttt{GPT-5.4-mini}~\citep{openai2026gpt54mini}, with the agent and world model sharing the same model. Additional implementation details and prompts are shown in Appendix~\ref{sec:appendix:implementation_details} and \ref{sec:appendix:prompts}.

\subsection{Experiments on World Model Prediction}
\label{sec:experiments:prediction}
Table~\ref{tab:stage2_main} evaluates next-observation prediction: (1) Among the baselines, RAWM-$\phi$ is strongest across \texttt{Gemma-4-26B}, \texttt{Qwen3.5-9B}, and \texttt{Gemma-4-31B}, showing that retrieval from collected trajectories provides useful transition evidence for next-observation prediction. By contrast, ITP-I consistently underperforms Zero-Shot, due to over-generation of imagined future details. (2) The memory ablations of \textsc{WorldEvolver} show complementary roles: Semantic Memory alone gives modest gains over Zero-Shot, whereas the Episodic Memory variant provides substantially larger improvements and outperforms RAWM-$\phi$, even though RAWM-$\phi$ retrieves from the full deployment trajectory set in advance while Episodic Memory accumulates strictly online. This gap suggests that retrieval quality depends on the retrieval key. RAWM-$\phi$ retrieves from full state-action text, where long and repetitive state descriptions can dilute the action signal, whereas episodic memory retrieval more directly matches the target transition being simulated. Representative cases in Appendix~\ref{sec:appendix:retrieval_failures} show that action-centered retrieval can still miss fine-grained executability constraints. The three metrics yield consistent rankings across backbones and environments, capturing correlated aspects of prediction quality in this evaluation.

Overall, accurate world model prediction benefits most from combining episodic retrieval with semantic rules. The full \textsc{WorldEvolver} achieves the highest prediction accuracy across both environments and all three backbones. Gains over episodic memory alone are particularly pronounced on ScienceWorld for the Gemma models, while Qwen3.5-9B shows smaller but consistent improvements from integrating semantic memory.

\begin{figure*}[t]
\centering
\includegraphics[width=\textwidth]{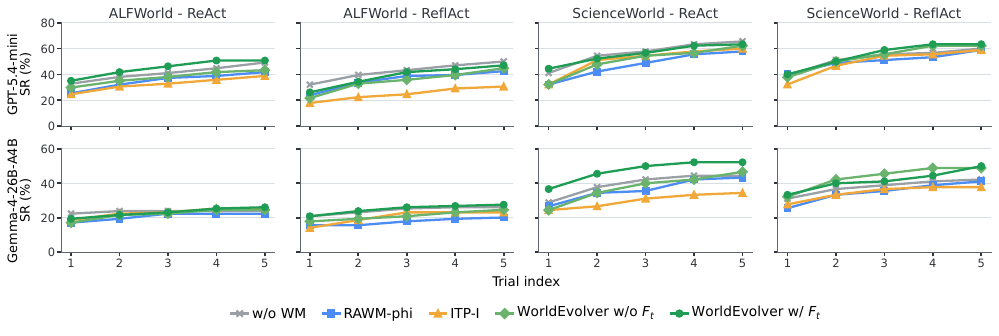}
\caption{Cumulative best-of-$L$ Agent Planning Success Rate on ALFWorld and ScienceWorld.}
\label{fig:pass_at_k_main}
\vspace{-0.3cm}
\end{figure*}

\begin{figure}[t]
\centering
\includegraphics[width=\columnwidth]{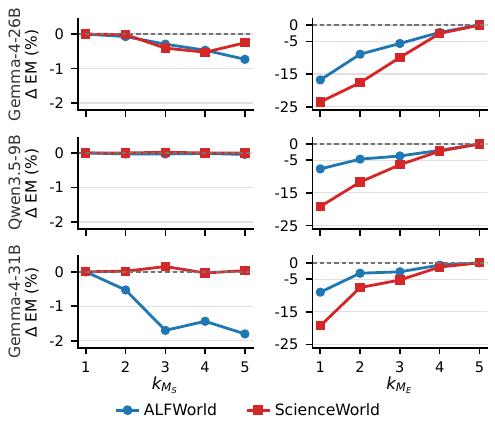}
\caption{Relative gains from memory hyperparameters for world model prediction, reported as $\Delta \text{EM}$. $k_{M_S}$ varies semantic-memory batch size relative to $1$, while $k_{M_E}$ varies episodic-memory retrieval size relative to $5$.}
\label{fig:prediction_memory_sweeps_exact_match}
\vspace{-0.3cm}
\end{figure}

\subsection{Experiments on Agent Planning}
\label{sec:experiments:agent_planning}

Table~\ref{tab:stage3_sr} evaluates agent planning by Success Rate: (1) Compared with world-model prediction results in Section~\ref{sec:experiments:prediction}, improving planning success is substantially more challenging. RAWM-$\phi$ and ITP-I often underperform the no-world-model baseline, further confirming that misaligned foresight can degrade action selection. (2) \textsc{WorldEvolver} is the strongest world-model method across all eight settings; selective foresight further improves or ties the no-foresight variant in every setting. Relative to RAWM-$\phi$, \textsc{WorldEvolver} w/o $F_t$ improves average Success Rate by 3.67 points, showing the advantage of continual episodic retrieval and mismatch-derived heuristic rule generation over static offline retrieval. (3) Improvements over the no-world-model baseline span both agent types and both model backbones. \textsc{WorldEvolver} w/o $F_t$ exceeds the no-world-model baseline in four settings: Gemma-4-26B-A4B with ReAct on ALFWorld (23.88 to 24.63), Gemma-4-26B-A4B on ScienceWorld for both ReAct and ReflAct (44.44 to 46.67; 42.22 to 48.89), and GPT-5.4-mini with ReflAct on ScienceWorld (60.00 to 62.22).

\textsc{WorldEvolver} w/ $F_t$ beats the no-world-model baseline across all four Gemma-4-26B-A4B cells, averaging $+2.24$ Success Rate points on ALFWorld and $+3.33$ on ScienceWorld over the no-foresight variant; even ReflAct on ALFWorld lifts \textsc{WorldEvolver} from $24.63$ to $27.61$, above the $26.12$ baseline. GPT-5.4-mini gains are more mixed: \textsc{WorldEvolver} w/ $F_t$ tops the no-world-model baseline in only two of four cells, winning by $+1.50$ on ReAct/ALFWorld and $+3.33$ on ReflAct/ScienceWorld but losing by $-2.99$ on ReflAct/ALFWorld and $-2.23$ on ReAct/ScienceWorld. Confidence-gated abstention is therefore more beneficial for the weaker backbone, where the agent leaves more room for useful world model guidance.

\subsection{Discussion}
\label{sec:discussion}
\paragraph{Memory Hyperparameters}
Figure~\ref{fig:prediction_memory_sweeps_exact_match} evaluates memory hyperparameters on the same setting as Table~\ref{tab:stage2_main}. Episodic retrieval is the dominant factor: increasing $k_{M_E}$ from 1 to 5 improves Exact Match by 16.8/23.5 points on ALFWorld and ScienceWorld for Gemma-4-26B-A4B, 7.6/19.2 for Qwen3.5-9B, and 9.0/19.3 for Gemma-4-31B. Semantic batch size is much less sensitive: most $k_{M_S}$ choices differ by within two points, except Gemma-4-31B on ALFWorld. We therefore use $k_{M_E}=5$ and $k_{M_S}=1$ in Section \ref{sec:experiments:prediction}, combining the strongest episodic retrieval with a semantic update size that is competitive across backbones and environments.

\paragraph{Online Continual Learning}
Figure~\ref{fig:pass_at_k_main} analyzes cumulative best-of-$L$ success rate from trial $L{=}1$ to $L{=}5$. The slope of each curve reflects the benefit of additional successful attempts beyond the first trial. This analysis is particularly relevant for \textsc{WorldEvolver}, since episodic memory $M_E$ and semantic memory $M_S$ accumulate across trials and tasks within the same environment, allowing later agent replanning to exploit refined world-model foresight. The clearest separation appears on ScienceWorld with Gemma-4-26B-A4B, where \textsc{WorldEvolver} variants increasingly outperform RAWM-$\phi$ and ITP-I as the trial index grows, demonstrating that deployment-time memory is most effective when agents can iteratively replan. Gains are smaller for GPT-5.4-mini because its stronger planning ability leaves less room for substantial improvement from world model foresight.

\paragraph{Memory Reset Controls}
To isolate cross-task accumulation from within-trajectory and within-task learning, we add reset controls for both $M_E$ and $M_S$. For World Model Prediction, both memories are cleared at the start of each trajectory. For Agent Planning, both memories are cleared before each task and retained across all five trials for that task. All control experiments use Gemma-4-26B-A4B; in Agent Planning, the agent and world model share this backbone. Tables~\ref{tab:memory_reset_prediction} and~\ref{tab:memory_reset_planning} report the corresponding results. The planning results are means over three complete runs, and the deltas use the reported full-memory \textsc{WorldEvolver} w/o $F_t$ results as the baseline for a direct comparison.

\begin{table}[t]
\centering
\small
\resizebox{\columnwidth}{!}{%
\begin{tabular}{@{}llrrr@{}}
\toprule
Environment & Memory setup & Exact Match & Token F1 & Cosine \\
\midrule
\multirow{2}{*}{ALFWorld}
& Full  & 0.5288 & 0.7675 & 0.8013 \\
& Reset & 0.4387 & 0.6675 & 0.7723 \\
\midrule
\multirow{2}{*}{ScienceWorld}
& Full  & 0.5155 & 0.6243 & 0.7385 \\
& Reset & 0.2070 & 0.3456 & 0.6178 \\
\bottomrule
\end{tabular}
}
\caption{World Model Prediction with persistent memory and memory reset before each trajectory.}
\label{tab:memory_reset_prediction}
\end{table}

\begin{table}[t]
\centering
\small
\resizebox{\columnwidth}{!}{%
\begin{tabular}{@{}llrrr@{}}
\toprule
Environment & Agent & Best-of-5 SR & Success@1 & $\Delta$ SR (pp) \\
\midrule
\multirow{2}{*}{ALFWorld}
& ReAct   & 24.1\% & 18.9\% & -0.5 \\
& ReflAct & 23.9\% & 19.7\% & -0.7 \\
\midrule
\multirow{2}{*}{ScienceWorld}
& ReAct   & 45.6\% & 23.3\% & -1.1 \\
& ReflAct & 47.8\% & 30.0\% & -1.1 \\
\bottomrule
\end{tabular}
}
\caption{Agent Planning with memory reset before each task, averaged over three runs. $\Delta$ SR is measured against full-memory \textsc{WorldEvolver} w/o $F_t$.}
\label{tab:memory_reset_planning}
\end{table}

Resetting memory substantially reduces next-observation prediction accuracy but changes best-of-5 planning success by only 0.5 to 1.1 points for two reasons. First, the evaluations measure different outcomes: World Model Prediction directly scores every one-step forecast, whereas Agent Planning scores final task completion. A wrong forecast therefore affects planning success only if it leads to a harmful action that the agent cannot correct after observing the actual outcome. Second, Agent Planning retains memory across the five trials of each task, so later trials can use evidence collected in earlier trials. Lower one-step accuracy therefore does not imply an equally large drop in best-of-5 success under this controlled reset protocol.

\paragraph{From Prediction to Planning}
Tables~\ref{tab:stage2_main} and~\ref{tab:stage3_sr} evaluate distinct stages of the pipeline. World Model Prediction directly scores static, one-step next-observation predictions, whereas Agent Planning evaluates their effect on closed-loop behavior. A prediction error matters only if it changes the agent's action and may then alter subsequent states. Improvements in one-step prediction accuracy therefore do not directly translate into higher terminal task success in closed-loop planning.

\begin{figure}[t]
\centering
\includegraphics[width=\columnwidth]{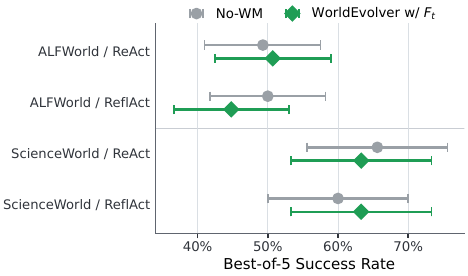}
\caption{GPT-5.4-mini Agent Planning best-of-5 Success Rate with task-level 95\% bootstrap confidence intervals across all four environment-agent settings.}
\label{fig:gpt_planning_ci}
\end{figure}

Compared with the smaller open-source Gemma-4-26B-A4B, GPT-5.4-mini has stronger general reasoning capabilities. Its parametric world knowledge and interaction history already support many action-effect inferences without an external world model, yielding a stronger no-world-model baseline and less headroom for additional foresight. Figure~\ref{fig:gpt_planning_ci} makes these mixed results explicit: the confidence intervals overlap with their no-world-model counterparts in all four settings.

\section{Conclusion}
\label{sec:conclusion}

We presented \textsc{WorldEvolver}, a training-free framework for self-evolving world models in LLM agent planning. Rather than updating model parameters, \textsc{WorldEvolver} revises world model context at test time through episodic memory, semantic memory, and selective foresight. Experiments on ALFWorld and ScienceWorld show that these mechanisms improve both world model fidelity and downstream planning performance. \textsc{WorldEvolver} achieves the strongest prediction accuracy on Word2World across three backbones and improves downstream agent success rates, suggesting that reliable foresight depends on how environmental signals are processed and presented to the agent, motivating future work on agentic world modeling.

\section*{Limitations}

\paragraph{Evaluation Scope}
To simplify evaluation and isolate the effects of deployment-time world-model revision from downstream agent behavior, we conduct experiments in two controlled long-horizon text environments, ALFWorld and ScienceWorld. This setting allows us to focus specifically on world-model \textit{foresight} and online adaptation, but does not cover broader domains such as web navigation, code generation, robotics, or multimodal interaction. Extending \textsc{WorldEvolver} to these settings is a natural direction for future work.

\paragraph{Confidence Estimation}
Our foresight filtering mechanism relies on prediction confidence signals derived from token-level probabilities, which may not be available in some closed-model APIs. In such settings, alternative confidence estimators, such as self-consistency or learned calibration models, would be required. In addition, the current dynamic filtering strategy assumes that prediction confidence correlates with prediction accuracy, as supported by Figure~\ref{fig:prediction_confidence_topx_em}, but this relationship can vary across environments and backbone models. We leave more robust confidence estimation and adaptive filtering mechanisms to future work.

\section*{Ethical Considerations}

Our code implementation is available at \url{https://github.com/magicgh/WorldEvolver}.
ALFWorld~\citep{shridhar2021alfworld}, ScienceWorld~\citep{wang2022scienceworld}, and the Word2World benchmark~\citep{li2025wordtoworld} are publicly available for research use. AI assistance was used as auxiliary support for coding and paper writing; all research decisions and claims are the authors' own.

\section*{Acknowledgments}

This research/project is supported by the National Research Foundation, Singapore under its National Large Language Models Funding Initiative (AISG Award No: AISG-NMLP-2024-002), and the Singapore Ministry of Education (MOE) Academic Research Fund (AcRF) Tier 1 grant (Proposal ID: 24-SIS-SMU-002). Any opinions, findings and conclusions or recommendations expressed in this material are those of the author(s) and do not reflect the views of National Research Foundation, Singapore.

\bibliography{custom}

\appendix
\section{Experimental Setups}
\label{sec:appendix:experimental_setups}

\paragraph{World Model Baselines}
We consider three inference-only baselines without gradient updates:
\begin{itemize}[leftmargin=*, itemsep=0.1em]
    \item \textbf{Zero-Shot} follows the standard zero-shot prompting paradigm for large language models~\citep{radford2019language}. The task description, current state, and proposed action are rendered directly as a next-observation prediction query.
    \item \textbf{RAWM-$\phi$}~\citep{yang2025rawm} reimplements RAWM as an offline retrieval baseline using only the retrieval encoder; $\phi$ denotes the absence of RAWM's PPO-trained MLP head, isolating the in-context retrieval contribution from the trained scoring component. It embeds the current query $(s_t,a_t)$ and stored transitions $(s_i,a_i,o_{i+1})$, retrieves the most similar top-1 transition by cosine similarity from a fixed retrieval library, and formats them as in-context examples for prediction. We use trajectories from the Word2World~\citep{li2025wordtoworld} test split as the retrieval source for both world-model prediction and agent planning. Retrieval is implemented with \texttt{Qwen3-Embedding-8B} for all retrieval queries across both evaluated environments under the shared embedding configuration.
    \item \textbf{ITP-I}~\citep{liu2026imaginethenplan} is the training-free variant of Imagine-then-Plan. In the original framework, adaptive lookahead operates within the agent planning loop, where the agent selects an imagination horizon and conditions action selection on the generated foresight. To isolate the effects of the world model while keeping the agent fixed, we move horizon selection and imagination into the world model itself. The world model selects $k \in \{0,\ldots,k_{\max}\}$ with $k_{\max}{=}5$ and returns the corresponding imagined future. For prediction evaluation, ITP-I is restricted to one-step imagination so that all methods share the same next-observation mismatch target. Multi-step imagination is used only in agent planning evaluation, where the downstream agent can consume longer horizon foresight during planning.
\end{itemize}

\paragraph{Agent Policies}
We evaluate two representative agent policies to test whether world-model signals transfer reliably across different agent types.
\begin{itemize}[leftmargin=*, itemsep=0.1em]
    \item \textbf{ReAct} follows the standard thought-action interaction format~\citep{yao2023react}, using in-context examples from the AgentBoard prompt.
    \item \textbf{ReflAct} augments ReAct with goal-state reflection before action selection~\citep{kim2025reflact}. This setting tests whether world-model predictions remain beneficial when the downstream agent already performs explicit reflection.
\end{itemize}

\section{Implementation Details}
\label{sec:appendix:implementation_details}
\begin{figure}[t]
  \centering
  \includegraphics[width=\columnwidth]{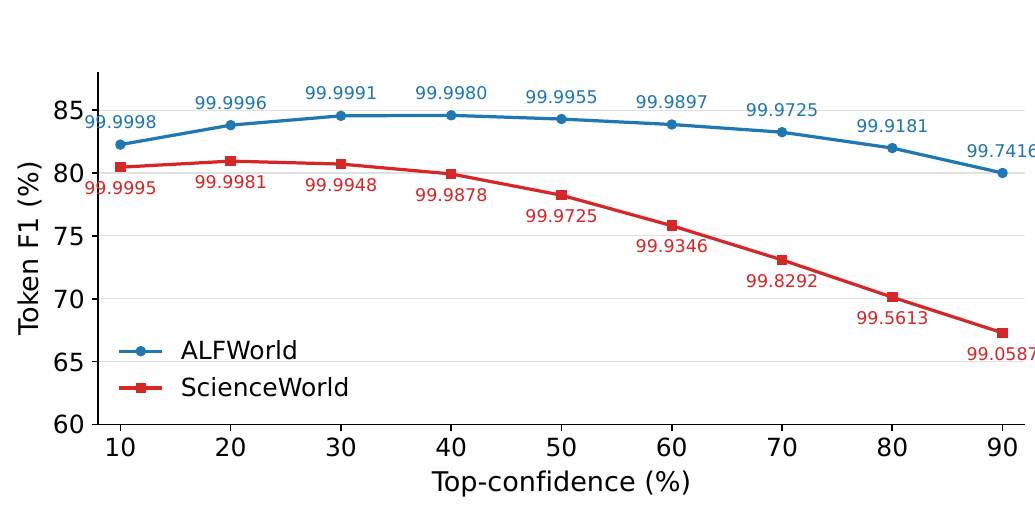}
  \caption{Selective foresight confidence calibration on Token F1 (\%) for Gemma-4-26B-A4B under the \textsc{WorldEvolver} configuration in both environments.}
  \label{fig:prediction_confidence_topx_f1_gemma}
\end{figure}

All generations use temperature $0$, top-$p$ sampling with $p{=}0.5$, random seed $42$, and a $32{,}768$-token context window. The mismatch critic and factorized-tuple mapping function $g$ share the same backbone as $W_\theta$. To support deployment-time continual learning, episodic memory $M_E$ and semantic memory $M_S$ persist across tasks within each environment. Selective foresight is applied when the geometric-mean token probability $q_t$ exceeds threshold $\tau$. Figures~\ref{fig:prediction_confidence_topx_f1_gemma} and~\ref{fig:prediction_confidence_topx_em} show that confidence scores correlate with both Exact Match and Token F1 on Gemma-4-26B-A4B. Per-cell thresholds (Table~\ref{tab:appendix_sf_thresholds}) are selected from these calibration curves, with values near $1{-}10^{-5}$ performing well; GPT-5.4-mini follows the same calibration procedure.

\begin{table}[t]
\centering
\scriptsize
\setlength{\tabcolsep}{3pt}
\resizebox{\columnwidth}{!}{%
\begin{tabular}{@{}lcccc@{}}
\toprule
\multirow{2}{*}{Agent} & \multicolumn{2}{c}{\textbf{Gemma-4-26B-A4B}} & \multicolumn{2}{c}{\textbf{GPT-5.4-mini}} \\
\cmidrule(lr){2-3}\cmidrule(l){4-5}
& ALFWorld & ScienceWorld & ALFWorld & ScienceWorld \\
\midrule
ReAct & $1-10^{-4}$ & $1-10^{-4}$ & $1-10^{-6}$ & $1-10^{-4}$ \\
ReflAct & $1-10^{-5}$ & $1-10^{-4}$ & $1-10^{-6}$ & $1-10^{-3}$ \\
\bottomrule
\end{tabular}
}
\caption{Values of $\tau$ used for selective foresight filtering in the w/ $F_t$ agent-planning settings across backbones.}
\label{tab:appendix_sf_thresholds}
\end{table}

\begin{table}[t]
\centering
\scriptsize
\setlength{\tabcolsep}{3pt}
\resizebox{\columnwidth}{!}{%
\begin{tabular}{@{}lrrrr@{}}
\toprule
\multirow{2}{*}{Method} & \multicolumn{3}{c}{Runtime (s/transition)} &  \\
\cmidrule(lr){2-4}
& ALFWorld & ScienceWorld & Average & GPU/h \\
\midrule
\multicolumn{5}{@{}c}{\textbf{Gemma-4-26B-A4B}} \\
Zero-Shot & 1.04 & 1.06 & 1.05 & 3.72 \\
RAWM-$\phi$ & 0.63 & 0.89 & 0.75 & 2.65 \\
ITP-I & 1.26 & 1.61 & 1.42 & 5.03 \\
\textsc{WorldEvolver} & 1.44 & 1.53 & 1.48 & 5.24 \\
\midrule
\multicolumn{5}{@{}c}{\textbf{Qwen3.5-9B}} \\
Zero-Shot & 0.53 & 0.80 & 0.66 & 2.32 \\
RAWM-$\phi$ & 0.56 & 0.91 & 0.72 & 2.55 \\
ITP-I & 1.15 & 1.33 & 1.23 & 4.36 \\
\textsc{WorldEvolver} & 0.69 & 1.27 & 0.96 & 3.38 \\
\midrule
\multicolumn{5}{@{}c}{\textbf{Gemma-4-31B}} \\
Zero-Shot & 0.70 & 0.82 & 0.75 & 2.66 \\
RAWM-$\phi$ & 0.51 & 0.53 & 0.52 & 1.84 \\
ITP-I & 0.97 & 1.45 & 1.19 & 4.21 \\
\textsc{WorldEvolver} & 1.25 & 1.45 & 1.34 & 4.73 \\
\bottomrule
\end{tabular}
}
\caption{World model prediction runtime, reported as seconds per evaluated transition and total GPU hours across both environments on a single NVIDIA H200 GPU. All methods use the same serving setup.}
\label{tab:appendix_stage2_runtime}
\end{table}

\section{Evaluation and Analysis}
\label{sec:appendix:evaluation_analysis}

\paragraph{Runtime}
The accuracy gains of \textsc{WorldEvolver} come from episodic and semantic memory modules, raising the question of whether these improvements justify the added inference cost. Table~\ref{tab:appendix_stage2_runtime} shows that \textsc{WorldEvolver} introduces only moderate runtime overhead relative to Zero-Shot and ITP-I. For example, on Gemma-4-26B-A4B, runtime increases from $1.05$s to $1.48$s per transition for \textsc{WorldEvolver}, compared to $1.42$s for ITP-I, while achieving substantially stronger Exact Match performance in Table~\ref{tab:stage2_main}. This suggests that the additional computation is effectively utilized for retrieval and mismatch-driven rule conditioning rather than longer imagination rollouts alone. RAWM-$\phi$ is the cheapest among most world-model approaches because retrieval embeddings are precomputed offline and excluded from runtime measurement. Despite this advantage, its prediction accuracy remains consistently below \textsc{WorldEvolver}. Overall, \textsc{WorldEvolver} provides the best trade-off between runtime and prediction performance across the evaluated backbones.

\paragraph{Memory Hyperparameter Analysis}
We first select $k_{M_E}{=}5$ using \textsc{WorldEvolver} w/o $M_S$ and $k_{M_S}{=}1$ using \textsc{WorldEvolver} w/o $M_E$. We then combine these values in the full model and use the same configuration across all environments and backbones instead of selecting separate hyperparameters for each setting. Tables~\ref{tab:appendix_kme_sweep} and~\ref{tab:appendix_kms_sweep} report the corresponding sweeps with Gemma-4-26B-A4B on ALFWorld and ScienceWorld, respectively.

\begin{table}[t]
\centering
\scriptsize
\setlength{\tabcolsep}{2.6pt}
\resizebox{\columnwidth}{!}{%
\begin{tabular}{@{}crrrcrrr@{}}
\toprule
\multirow{2}{*}{$k_{M_E}$} & \multicolumn{3}{c}{ALFWorld} && \multicolumn{3}{c}{ScienceWorld} \\
\cmidrule(lr){2-4}\cmidrule(l){6-8}
& EM & F1 & Cosine && EM & F1 & Cosine \\
\midrule
1 & 0.4006 & 0.6931 & 0.7741 && 0.1585 & 0.3098 & 0.5888 \\
2 & 0.4817 & 0.7331 & 0.7909 && 0.3624 & 0.4922 & 0.6834 \\
3 & 0.4933 & 0.7383 & 0.7925 && 0.4764 & 0.5960 & 0.7297 \\
4 & 0.5253 & 0.7623 & 0.7997 && 0.4811 & 0.5961 & 0.7275 \\
5 & 0.5290 & 0.7646 & 0.7977 && 0.5038 & 0.6064 & 0.7312 \\
\bottomrule
\end{tabular}
}
\caption{Varying Episodic Memory retrieval size $k_{M_E}$ with $k_{M_S}{=}1$. The $k_{M_E}{=}5$ row reports the five-seed aggregate; the other rows report canonical runs.}
\label{tab:appendix_kme_sweep}
\end{table}

\begin{table}[t]
\centering
\scriptsize
\setlength{\tabcolsep}{2.6pt}
\resizebox{\columnwidth}{!}{%
\begin{tabular}{@{}crrrcrrr@{}}
\toprule
\multirow{2}{*}{$k_{M_S}$} & \multicolumn{3}{c}{ALFWorld} && \multicolumn{3}{c}{ScienceWorld} \\
\cmidrule(lr){2-4}\cmidrule(l){6-8}
& EM & F1 & Cosine && EM & F1 & Cosine \\
\midrule
1 & 0.5288 & 0.7675 & 0.8013 && 0.5155 & 0.6243 & 0.7385 \\
2 & 0.5313 & 0.7662 & 0.8007 && 0.4772 & 0.5889 & 0.7223 \\
3 & 0.5334 & 0.7701 & 0.8018 && 0.5540 & 0.6573 & 0.7565 \\
4 & 0.5379 & 0.7762 & 0.8023 && 0.4909 & 0.6034 & 0.7308 \\
5 & 0.5388 & 0.7776 & 0.8028 && 0.4994 & 0.6134 & 0.7348 \\
\bottomrule
\end{tabular}
}
\caption{Semantic Memory batch-size sweep ($k_{M_E}{=}5$).}
\label{tab:appendix_kms_sweep}
\end{table}

The best tested combination is $(k_{M_E},k_{M_S}){=}(5,5)$ on ALFWorld and $(5,3)$ on ScienceWorld. We nevertheless use the shared setting $(5,1)$ across environments and backbones. This configuration is selected on Word2World and transferred unchanged to AgentBoard; therefore, the memory hyperparameters are not tuned to the AgentBoard planning results in this evaluation.

\begin{figure*}[t]
  \centering
  \includegraphics[width=\textwidth]{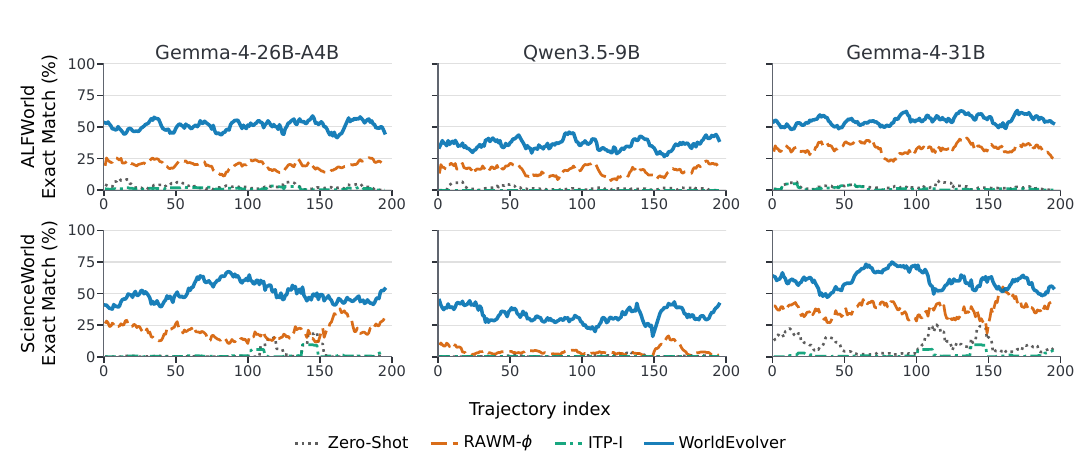}
  \caption{Trajectory-level world model prediction Exact Match (\%) along the Word2World deployment order. We report macro Exact Match averaged over the prediction steps within each trajectory.}
  \label{fig:appendix_stage2_trajectory_em}
\end{figure*}

\paragraph{Memory Evolution}
Figure~\ref{fig:appendix_stage2_trajectory_em} plots trajectory-macro Exact Match in deployment order. \textsc{WorldEvolver} consistently stays in a higher accuracy band than Zero-Shot, RAWM-$\phi$, and ITP-I across environments and backbones, indicating that online memories provide reusable context beyond the current trajectory. The separation is most pronounced for the Gemma family models. On ScienceWorld, \textsc{WorldEvolver} shows a clear mid-deployment lift, while on ALFWorld it remains high and stable throughout, suggesting that ALFWorld's more regular transition structure enables earlier reuse of accumulated evidence. Qwen3.5-9B exhibits lower and noisier local Exact Match, implying that memory evidence is less effective when the backbone model is less reliable at predicting next observations during deployment.

\begin{figure}[t]
  \centering
  \includegraphics[width=\columnwidth]{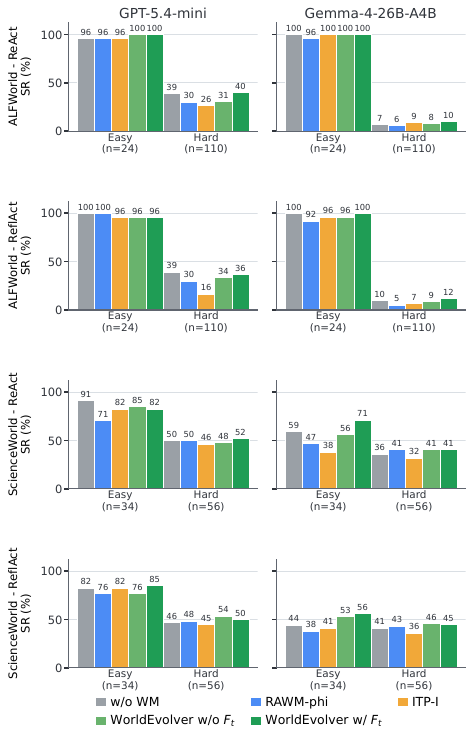}
  \caption{Agent Planning Success Rate (\%) by AgentBoard easy/hard split, reported as best-of-5.}
  \label{fig:appendix_stage3_sr_by_difficulty}
\end{figure}

\paragraph{Difficulty Breakdown}
Figure~\ref{fig:appendix_stage3_sr_by_difficulty} reports success rate by task difficulty. ALFWorld easy tasks are nearly saturated for both backbones, making the hard split more informative. On the hard version of ALFWorld, GPT-5.4-mini already achieves substantially higher success than Gemma-4-26B-A4B, leaving limited room for additional foresight gains; for Gemma-4-26B-A4B, \textsc{WorldEvolver} yields small improvements mainly when selective foresight is enabled. ScienceWorld is less saturated, especially for Gemma-4-26B-A4B, so differences among world-model methods are more visible. In this setting, \textsc{WorldEvolver} improves Gemma-4-26B-A4B across both agent types and gives the clearest GPT-5.4-mini gain on ReflAct hard tasks, increasing success from $46.00$ to $54.00$ without $F_t$ and $50.00$ with $F_t$. Overall, world model foresight is most useful when tasks are not saturated and transition uncertainty remains across environments.

\begin{figure*}[t]
  \centering
  \includegraphics[width=0.95\textwidth]{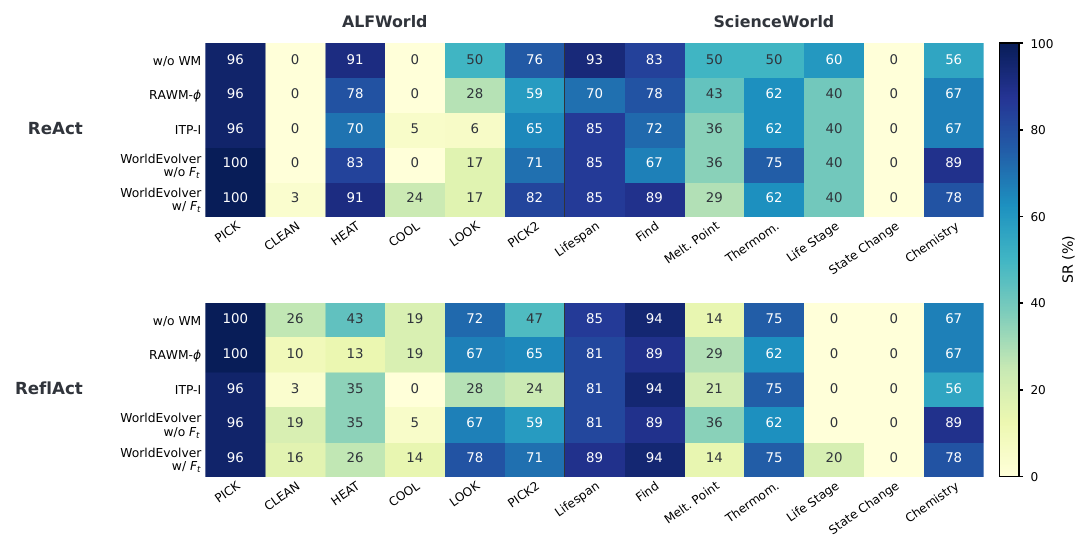}
  \caption{Heatmap of agent planning success rates (\%) for GPT-5.4-mini across different world models, reported as best-of-5 performance on ALFWorld and ScienceWorld task types.}
  \label{fig:appendix_stage3_sr_by_task_type_heatmap}
\end{figure*}

\begin{figure*}[t]
  \centering
  \includegraphics[width=0.95\textwidth]{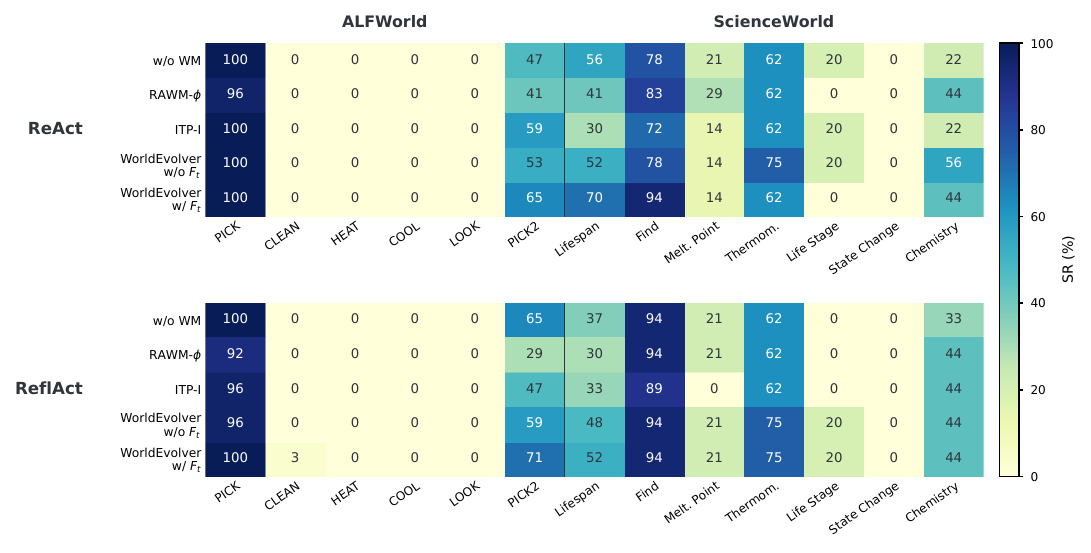}
  \caption{Heatmap of agent planning success rates (\%) for Gemma-4-26B-A4B across different world models, reported as best-of-5 performance on ALFWorld and ScienceWorld task types.}
  \label{fig:appendix_stage3_sr_by_task_type_heatmap_gemma}
\end{figure*}

\paragraph{Task Type Breakdown}
Figures~\ref{fig:appendix_stage3_sr_by_task_type_heatmap} and~\ref{fig:appendix_stage3_sr_by_task_type_heatmap_gemma} depict the success rates of GPT-5.4-mini and Gemma-4-26B-A4B by task type. The heatmaps show a consistent pattern across backbones. On ALFWorld, \texttt{PICK} is nearly saturated, while \texttt{CLEAN}, \texttt{COOL}, and \texttt{LOOK} remain difficult, especially for Gemma-4-26B-A4B. The clearest gains appear on transition-sensitive types such as \texttt{PICK2}, where \textsc{WorldEvolver} improves both backbones and both agent policies, suggesting that weaker planners leave more room for useful foresight. On ScienceWorld, gains concentrate on task families that require tracking environment dynamics, including \texttt{Lifespan}, \texttt{Thermom.}, and \texttt{Chemistry}, while \texttt{State Change} remains near zero across all methods and backbones. This suggests that world model foresight captures reusable task-family dynamics, but remains limited when relevant transitions are too sparse to be reliably accumulated by $M_E$ or abstracted into $M_S$ for reuse across subsequent tasks.

\paragraph{Action-Level Effect of Selective Foresight}
For Gemma-4-26B-A4B, we measure how often visible selective foresight changes the draft action, aggregated over all foresight-exposed decision steps in the AgentBoard planning rollouts. All rows in Table~\ref{tab:foresight_action_changes} correspond to the w/ $F_t$ condition.

\begin{table}[t]
\centering
\small
\resizebox{\columnwidth}{!}{%
\begin{tabular}{@{}llrrr@{}}
\toprule
Environment & Agent & Steps with $F_t$ & Action changed & Change rate \\
\midrule
\multirow{2}{*}{ALFWorld}
& ReAct   & 17,867 & 1,614 & 9.0\% \\
& ReflAct & 17,981 & 946   & 5.3\% \\
\midrule
\multirow{2}{*}{ScienceWorld}
& ReAct   & 11,450 & 1,912 & 16.7\% \\
& ReflAct & 10,981 & 1,376 & 12.5\% \\
\bottomrule
\end{tabular}
}
\caption{Action changes after selective foresight is shown to Gemma-4-26B-A4B during Agent Planning.}
\label{tab:foresight_action_changes}
\end{table}

The world model adds $F_t$ to the agent's interaction history, which the agent may use to revise its draft action; this occurs on 5.3\% to 16.7\% of exposed steps. Figures~\ref{fig:appendix_stage3_sr_by_task_type_heatmap} and~\ref{fig:appendix_stage3_sr_by_task_type_heatmap_gemma} compare \textsc{WorldEvolver} with the no-world-model baseline by task type. The clearest gains occur on transition-sensitive ALFWorld \texttt{PICK2} tasks and ScienceWorld \texttt{Lifespan}, \texttt{Thermom.}, and \texttt{Chemistry} tasks, whereas ALFWorld \texttt{PICK} is nearly saturated and ScienceWorld \texttt{State Change} remains near zero across methods. These results identify the task types on which foresight is associated with higher task success across the evaluated benchmarks.

\begin{figure*}[t]
\centering
\includegraphics[width=\textwidth]{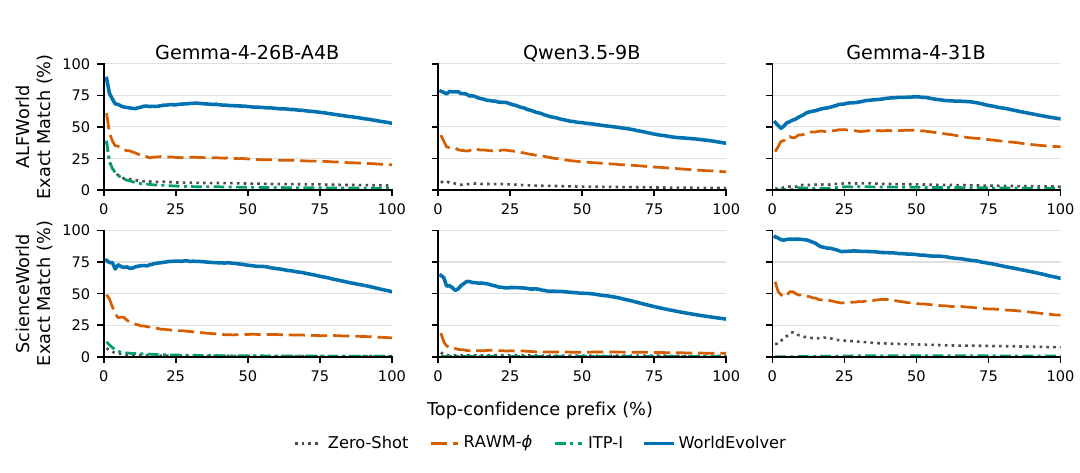}
\caption{Selective foresight confidence calibration, measured as Exact Match on the top-confidence prefix.}
\label{fig:prediction_confidence_topx_em}
\end{figure*}

\paragraph{Foresight Confidence}
Figure~\ref{fig:prediction_confidence_topx_em} reports Exact Match over predictions ranked by \textsc{WorldEvolver}'s confidence across quantiles. In most settings, Exact Match decreases as confidence coverage expands, indicating that higher-confidence predictions are generally more reliable and can support selective foresight. Across all retention percentages, \textsc{WorldEvolver} remains well above Zero-Shot, RAWM-$\phi$, and ITP-I, suggesting that the confidence gate filters a stronger predictive signal rather than merely selecting examples that are easy for all methods. This aligns with Table~\ref{tab:stage3_sr}, where adding $F_t$ consistently matches or improves over removing $F_t$ across planning settings.

AgentBoard reports two planning metrics alongside Success Rate. \textbf{Progress Rate} records the maximum task progress reached across trials. \textbf{Grounding} measures how often the environment accepts the agent's actions as valid. Table~\ref{tab:stage3_aux_appendix} reports both metrics for every cell of Table~\ref{tab:stage3_sr} for completeness.

\begin{table*}[t]
\centering
\tiny
\resizebox{\textwidth}{!}{%
\begin{tabular}{@{}llcccc@{}}
\toprule
\multirow{2}{*}{Agent Mode} & \multirow{2}{*}{Condition} & \multicolumn{2}{c}{ALFWorld} & \multicolumn{2}{c}{ScienceWorld} \\
\cmidrule(lr){3-4}\cmidrule(l){5-6}
& & Progress Rate & Grounding & Progress Rate & Grounding \\
\midrule
\multicolumn{6}{@{}c}{\textbf{Gemma-4-26B-A4B}} \\
\midrule
\multirow{4}{*}{ReAct}
& w/o World Model & 61.57 & 72.13 & 76.58 & 35.76 \\
& RAWM-$\phi$ \cite{yang2025rawm} & 60.70 & 76.77 & 74.88 & 34.28 \\
& ITP-I \cite{liu2026imaginethenplan} & 63.25 & 75.17 & 73.42 & 40.31 \\
& \textsc{WorldEvolver} w/o $F_t$ & 61.57 & 79.76 & 79.47 & 35.02 \\
\midrule
\multirow{4}{*}{ReflAct}
& w/o World Model & 63.81 & 70.25 & 75.22 & 33.09 \\
& RAWM-$\phi$ \cite{yang2025rawm} & 58.27 & 69.29 & 73.89 & 29.41 \\
& ITP-I \cite{liu2026imaginethenplan} & 61.38 & 72.95 & 74.68 & 36.65 \\
& \textsc{WorldEvolver} w/o $F_t$ & 62.38 & 77.30 & 77.58 & 26.93 \\
\midrule
\multicolumn{6}{@{}c}{\textbf{GPT-5.4-mini}} \\
\midrule
\multirow{4}{*}{ReAct}
& w/o World Model & 75.31 & 76.05 & 88.31 & 38.62 \\
& RAWM-$\phi$ \cite{yang2025rawm} & 73.82 & 79.62 & 84.74 & 40.23 \\
& ITP-I \cite{liu2026imaginethenplan} & 70.52 & 83.61 & 85.62 & 40.40 \\
& \textsc{WorldEvolver} w/o $F_t$ & 74.07 & 77.04 & 87.39 & 40.59 \\
\midrule
\multirow{4}{*}{ReflAct}
& w/o World Model & 76.12 & 78.81 & 84.43 & 40.89 \\
& RAWM-$\phi$ \cite{yang2025rawm} & 71.27 & 78.21 & 83.25 & 39.07 \\
& ITP-I \cite{liu2026imaginethenplan} & 66.23 & 78.14 & 82.87 & 42.79 \\
& \textsc{WorldEvolver} w/o $F_t$ & 71.95 & 82.23 & 82.69 & 41.56 \\
\bottomrule
\end{tabular}
}
\caption{AgentBoard planning metrics for the Success Rate results in Table~\ref{tab:stage3_sr}. Values are percentages without percent signs for both backbones and environments.}
\label{tab:stage3_aux_appendix}
\end{table*}
\paragraph{Comparison of Retrieval Methods}
We compare the top-1 transition-retrieval accuracy of \textsc{WorldEvolver}'s action-based Episodic Memory and RAWM-$\phi$~\citep{yang2025rawm}, using Gemma-4-26B-A4B as the shared backbone throughout.

\begin{table}[t]
\centering
\small
\begin{tabular*}{\columnwidth}{@{\extracolsep{\fill}}lrrr@{}}
\toprule
Environment & Action-Jaccard & RAWM-$\phi$ & $\Delta$ \\
\midrule
ALFWorld & 0.783 & 0.529 & +0.254 \\
ScienceWorld & 0.687 & 0.653 & +0.034 \\
\bottomrule
\end{tabular*}
\caption{Top-1 transition-retrieval accuracy. $\Delta$ denotes Action-Jaccard minus RAWM-$\phi$ across environments.}
\label{tab:appendix_retrieval_accuracy}
\end{table}

As discussed in Section~\ref{sec:experiments:prediction}, action-centered retrieval outperforms full state-action retrieval in both environments, with a particularly large gain on ALFWorld. State descriptions in these environments are lengthy and highly repetitive, whereas action strings already encode task-relevant state information, including object identities and relations, as in \textit{``take soapbottle 1 from shelf 1''} in ALFWorld and \textit{``connect battery cathode to green light bulb cathode''} in ScienceWorld. RAWM-$\phi$ performs $(s,a)\rightarrow(s,a,s')$ retrieval: it matches the full current state-action pair $(s_t,a_t)$ against stored state-action keys $(s_i,a_i)$ and returns the associated transition $(s_i,a_i,s_{i+1})$. Repetitive state descriptions can therefore dominate the comparison and dilute the action signal during transition matching.

\begin{figure}[t]
\centering
\includegraphics[width=\columnwidth]{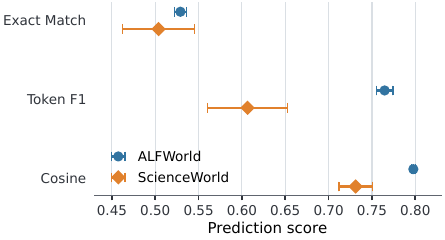}
\caption{World Model Prediction across five randomly seeded trajectory orders with uncertainty bands.}
\label{fig:prediction_order_robustness}
\end{figure}

\paragraph{Robustness to Evaluation Order}
We also evaluate five random seeds for World Model Prediction, each defining a different evaluation order over trajectories. Figure~\ref{fig:prediction_order_robustness} reports the mean and standard deviation across the five orders in each environment under the shared evaluation protocol used here.

The results are consistent across the five seeded trajectory orders, indicating that performance is not sensitive to the specific order in either environment.

\paragraph{Statistical Uncertainty and Success@1}
For Gemma-4-26B-A4B, we compute task-level 95\% bootstrap confidence intervals from 10,000 resamples and report Success@1, which records task success on the first trial. Each cell in Table~\ref{tab:appendix_stage3_bootstrap_success1} reports best-of-5 Success Rate [95\% CI] / Success@1 in percentage points for direct comparison.

\begin{table}[t]
\centering
\scriptsize
\setlength{\tabcolsep}{2pt}
\resizebox{\columnwidth}{!}{%
\begin{tabular}{@{}lcc@{}}
\toprule
\multicolumn{3}{c}{\textbf{ALFWorld}} \\
\midrule
Method & ReAct & ReflAct \\
\midrule
w/o World Model & $23.9\ [17,31]\,/\,22.4$ & $26.1\ [19,34]\,/\,20.9$ \\
RAWM-$\phi$ & $22.4\ [16,29]\,/\,17.2$ & $20.1\ [14,27]\,/\,15.7$ \\
ITP-I & $25.4\ [19,33]\,/\,18.7$ & $23.1\ [16,31]\,/\,14.2$ \\
\textsc{WorldEvolver} w/o $F_t$ & $24.6\ [18,32]\,/\,17.2$ & $24.6\ [18,32]\,/\,17.9$ \\
\textsc{WorldEvolver} w/ $F_t$ & $26.1\ [19,34]\,/\,19.4$ & $27.6\ [21,35]\,/\,20.9$ \\
\midrule
\multicolumn{3}{c}{\textbf{ScienceWorld}} \\
\midrule
Method & ReAct & ReflAct \\
\midrule
w/o World Model & $44.4\ [34,56]\,/\,28.9$ & $42.2\ [32,52]\,/\,31.1$ \\
RAWM-$\phi$ & $43.3\ [33,53]\,/\,26.7$ & $41.1\ [31,51]\,/\,25.6$ \\
ITP-I & $34.4\ [24,44]\,/\,24.4$ & $37.8\ [28,48]\,/\,27.8$ \\
\textsc{WorldEvolver} w/o $F_t$ & $46.7\ [37,57]\,/\,24.4$ & $48.9\ [39,59]\,/\,32.2$ \\
\textsc{WorldEvolver} w/ $F_t$ & $52.2\ [42,62]\,/\,36.7$ & $50.0\ [40,60]\,/\,33.3$ \\
\bottomrule
\end{tabular}
}
\caption{Agent Planning uncertainty and first-trial success for Gemma-4-26B-A4B. Each cell reports best-of-5 Success Rate [task-level 95\% bootstrap CI] / Success@1 (\%) across all four environment-agent settings.}
\label{tab:appendix_stage3_bootstrap_success1}
\end{table}

\textsc{WorldEvolver} w/ $F_t$ has the highest best-of-5 point estimate in all four settings, although the confidence intervals overlap. It also achieves the highest Success@1 among world-model methods in all four settings. Across all methods, its Success@1 is highest in both ScienceWorld settings, ties the no-world-model baseline on ALFWorld ReflAct, and trails it on ALFWorld ReAct.

\paragraph{Planning-Time Overhead and Memory Growth}
Table~\ref{tab:appendix_stage3_overhead} reports \textsc{WorldEvolver} w/o $F_t$ using Gemma-4-26B-A4B with $k_{M_E}{=}5$ and $k_{M_S}{=}1$. WM invocations include prediction and memory-update calls, whereas LLM calls include all underlying LLM calls. Maximum prompt tokens measure the peak context length associated with memory growth during closed-loop planning.

\begin{table}[t]
\centering
\scriptsize
\setlength{\tabcolsep}{2pt}
\resizebox{\columnwidth}{!}{%
\begin{tabular}{@{}lrrrr@{}}
\toprule
Setting & \shortstack{WM invocations \\ / agent step} & \shortstack{LLM calls \\ / agent step} & \shortstack{Retrieval \\ ms / call} & \shortstack{Max prompt tokens \\ / WM call} \\
\midrule
ALFWorld ReAct & 2.086 & 1.609 & 16.244 & 2,982 \\
ALFWorld ReflAct & 2.054 & 1.642 & 18.335 & 2,726 \\
ScienceWorld ReAct & 2.157 & 1.866 & 10.100 & 4,324 \\
ScienceWorld ReflAct & 2.131 & 1.874 & 10.906 & 4,211 \\
\bottomrule
\end{tabular}
}
\caption{Planning-time overhead and peak context length for \textsc{WorldEvolver} w/o $F_t$ during planning.}
\label{tab:appendix_stage3_overhead}
\end{table}

The two representative failures in Section~\ref{sec:appendix:retrieval_failures} show how relevant memory can still mislead prediction. In ALFWorld, retrieved successful pickup transitions cause the model to predict a pickup for \textit{``take soapbottle 2 from garbagecan 1,''} although the gold observation is \textit{``Nothing happens.''} In ScienceWorld, same-task circuit transitions cause the model to predict a successful connection for \textit{``connect green light bulb anode to switch cathode using yellow wire,''} although the simulator returns \textit{``No known action matches that input.''} In both cases, retrieval identifies the correct task domain but misses a fine-grained executability or command-validity constraint during action-conditioned prediction.

\paragraph{Selective-Foresight Threshold Calibration}
Selective Foresight is used only in Agent Planning. As detailed in Appendix~\ref{sec:appendix:implementation_details}, thresholds are selected from the Word2World test-set confidence-quality curves in Figure~\ref{fig:prediction_confidence_topx_f1_gemma}, which sweep confidence-retention levels and report Token F1. These curves provide prediction-level threshold sensitivity. We then fix the selected thresholds and transfer them unchanged to AgentBoard, so no AgentBoard planning outcome is used for threshold selection.

To reduce sensitivity to a single evaluation order, we aggregate post-hoc calibration diagnostics across five randomly seeded trajectory orders for World Model Prediction in each environment.

\begin{table}[t]
\centering
\scriptsize
\setlength{\tabcolsep}{2pt}
\resizebox{\columnwidth}{!}{%
\begin{tabular}{@{}lrrrrr@{}}
\toprule
Environment & \shortstack{Predictions \\ scored} & Mean F1 & \shortstack{Mean \\ confidence} & \shortstack{ECE-F1-10 \\ $\downarrow$} & \shortstack{F1 C-index \\ $\uparrow$} \\
\midrule
ALFWorld & 34,150 & 0.7646 & 0.9912 & 0.2266 & 0.6632 \\
ScienceWorld & 29,489 & 0.6065 & 0.9771 & 0.3706 & 0.6963 \\
\bottomrule
\end{tabular}
}
\caption{Confidence calibration and ranking diagnostics across five seeded trajectory orders in aggregate.}
\label{tab:appendix_confidence_calibration}
\end{table}

Selective Foresight uses $q_t$ to rank and filter predictions rather than as a probability estimate. ECE-F1-10 is the sample-weighted absolute gap between mean confidence and mean Token F1 across ten confidence bins. Because Token F1 is our primary prediction-quality metric and is continuous, we report C-index for ranking instead of ROC-AUC, which would require binarizing F1 at an arbitrary threshold. C-index measures how often higher confidence corresponds to higher F1 across comparable pairs, with 0.5 indicating random ranking and 1.0 perfect ranking. Scores of 0.6632 and 0.6963 indicate 66.3\% and 69.6\% correct ordering on ALFWorld and ScienceWorld, respectively.

\section{Case Study}
\label{sec:appendix:retrieval_failures}

We present one retrieval failure from each environment. In both cases, Episodic Memory retrieves task-relevant transitions, but the model fails to recognize that the specific action cannot be executed and incorrectly predicts a successful outcome.

\begin{table}[h]
\centering
\scriptsize
\setlength{\tabcolsep}{3pt}
\begin{tabular}{@{}p{0.22\columnwidth}p{0.70\columnwidth}@{}}
\toprule
Field & Value \\
\midrule
Task & put two soapbottle in cart \\
Target action & take soapbottle 2 from garbagecan 1 \\
Model prediction & You pick up the soapbottle 2 from the garbagecan 1. \\
Gold next state & Nothing happens. \\
\bottomrule
\end{tabular}

\vspace{0.5em}

\begin{tabular}{@{}p{0.08\columnwidth}p{0.34\columnwidth}p{0.50\columnwidth}@{}}
\toprule
Rank & Source action & Source gold \\
\midrule
1 & take soapbottle 1 from shelf 1 & You pick up the soapbottle 1 from the shelf 1. \\
2 & go to garbagecan 1 & You arrive at garbagecan 1. On the garbagecan 1, you see a papertowelroll 1, and a soapbottle 2. \\
3 & examine sinkbasin 1 & On the sinkbasin 1, you see a soapbar 1. \\
4 & go to shelf 1 & You arrive at shelf 1. On the shelf 1, you see a soapbottle 1. \\
5 & go to shelf 2 & You arrive at shelf 2. On the shelf 2, you see nothing. \\
\bottomrule
\end{tabular}
\caption{Representative Episodic Memory retrieval failure in ALFWorld with retrieved transitions.}
\label{tab:appendix_retrieval_failure_alfworld}
\end{table}

Retrieval surfaces closely related same-task transitions involving the relevant object class and location. The failure arises because the current transition depends on a fine-grained executability condition: although similar pickup and location memories are relevant, the specific target command produces no state change. The model therefore overgeneralizes from nearby successful transitions and predicts a pickup despite the no-op outcome.

\begin{table}[h]
\centering
\scriptsize
\setlength{\tabcolsep}{3pt}
\begin{tabular}{@{}p{0.22\columnwidth}p{0.70\columnwidth}@{}}
\toprule
Field & Value \\
\midrule
Task & electrical conductivity of unknown substance L \\
Target action & connect green light bulb anode to switch cathode using yellow wire \\
Model prediction & The yellow wire connects green bulb anode to switch cathode. \\
Gold next state & No known action matches that input. \\
\bottomrule
\end{tabular}

\vspace{0.5em}

\begin{tabular}{@{}p{0.08\columnwidth}p{0.34\columnwidth}p{0.50\columnwidth}@{}}
\toprule
Rank & Source action & Source gold \\
\midrule
1 & connect battery cathode to green light bulb & Connections must specify specific points of contact on polarized electrical components\ldots \\
2 & connect battery cathode to green light bulb cathode & cathode on battery is now connected to cathode on green light bulb \\
3 & connect battery anode to switch & Connections must specify specific points of contact on polarized electrical components\ldots \\
4 & connect battery anode to switch anode & anode on battery is now connected to anode on switch \\
5 & connect battery to switch & Connections must specify specific points of contact on polarized electrical components\ldots \\
\bottomrule
\end{tabular}
\caption{Representative Episodic Memory retrieval failure in ScienceWorld with circuit transitions.}
\label{tab:appendix_retrieval_failure_scienceworld}
\end{table}

Retrieval is domain-relevant: all retrieved transitions involve circuit construction in the same task family. The failure is that simulator command validity is highly specific. Similar connection memories help identify the correct physical domain, but they do not guarantee that every phrasing of a connection command is accepted by the environment. The model overgeneralizes from nearby valid circuit updates and predicts a successful connection. Together, these cases show that action-centered retrieval can identify the relevant domain while still missing fine-grained executability constraints.

\section{Prompts}
\label{sec:appendix:prompts}

The ReAct and ReflAct prompts for ALFWorld are shown in Figures~\ref{fig:prompt_agent_react} and~\ref{fig:prompt_agent_reflact}, with ScienceWorld counterparts in Figures~\ref{fig:prompt_agent_react_sw} and~\ref{fig:prompt_agent_reflact_sw}. World-model prompts are provided in Figures~\ref{fig:prompt_wm_zeroshot}--\ref{fig:prompt_wm_worldevolver}. The two memory-update prompts used in \textsc{WorldEvolver} are shown in Figures~\ref{fig:prompt_factorizer} and~\ref{fig:prompt_rule_extractor}.

\label{sec:appendix:prompts:agent}

\begin{figure*}[t]
\centering
\begin{tcolorbox}[
    title=\textbf{Prompt Template for ReAct Agent (ALFWorld)},
    colframe=blue!50!black, colback=blue!5!white, coltitle=white,
    fonttitle=\bfseries, arc=1mm, boxsep=2pt, fontupper=\small,
]
Your task is to interact with a virtual household simulator to accomplish a specific task. With each interaction, you will receive an observation. Your role is to decide on an action based on the observation. Please ensure that any objects (\texttt{\{obj\}}) and receptacles (\texttt{\{recep\}}) you mention in your response are present in the observation provided.

\vspace{4pt}
\textbf{Available actions:}
\begin{tcolorbox}[colback=white, colframe=gray!30, arc=0mm, boxsep=0pt, left=4pt, right=4pt, top=2pt, bottom=2pt]
\ttfamily\footnotesize
take \{obj\} from \{recep\}\;\;put \{obj\} in/on \{recep\}\;\;open \{recep\}\;\;close \{recep\}\;\;toggle \{obj\}/\{recep\}\;\;clean \{obj\} using \{recep\}\;\;cool \{obj\} using \{recep\}\;\;heat \{obj\} using \{recep\}\;\;inventory\;\;examine \{recep\}/\{obj\}\;\;go to \{recep\}
\end{tcolorbox}

\vspace{4pt}
For each of your turns, you will be given the observation of the last turn. You should first think about what to do, and then output the action for this turn.

\vspace{4pt}
\textbf{Response format:}
\begin{tcolorbox}[colback=white, colframe=gray!30, arc=0mm, boxsep=0pt, left=4pt, right=4pt, top=2pt, bottom=2pt]
\ttfamily\footnotesize
Thought: \textless your thought\textgreater\par
Action: \textless your next action\textgreater
\end{tcolorbox}
Return exactly one line. Do not begin with a newline, blank line, space, markdown, or any text before ``Thought:'' or ``Action:''.
\end{tcolorbox}
\caption{ReAct agent prompt for ALFWorld, with placeholders for available actions and the response format.}
\label{fig:prompt_agent_react}
\end{figure*}

\begin{figure*}[t]
\centering
\begin{tcolorbox}[
    title=\textbf{Prompt Template for ReflAct Agent (ALFWorld)},
    colframe=blue!50!black, colback=blue!5!white, coltitle=white,
    fonttitle=\bfseries, arc=1mm, boxsep=2pt, fontupper=\small,
]
Your task is to interact with a virtual household simulator to accomplish a specific task. With each interaction, you will receive an observation. Your role is to decide on an action based on the observation. The available actions and object/receptacle constraints match the ReAct prompt above (Figure~\ref{fig:prompt_agent_react}).

\vspace{4pt}
For each of your turns, you should first reflect in one sentence on the agent's state in relation to the task goal, and then output the action for this turn.

\vspace{4pt}
\textbf{Response format:}
\begin{tcolorbox}[colback=white, colframe=gray!30, arc=0mm, boxsep=0pt, left=4pt, right=4pt, top=2pt, bottom=2pt]
\ttfamily\footnotesize
Reflection: \textless your reflection\textgreater\par
Action: \textless your next action\textgreater
\end{tcolorbox}
Return exactly one line. Do not begin with a newline, blank line, space, markdown, or any text before ``Reflection:'' or ``Action:''.
\end{tcolorbox}
\caption{ReflAct agent prompt for ALFWorld, replacing the ReAct \texttt{Thought} with \texttt{Reflection}.}
\label{fig:prompt_agent_reflact}
\end{figure*}

\begin{figure*}[t]
\centering
\begin{tcolorbox}[
    title=\textbf{Prompt Template for ReAct Agent (ScienceWorld)},
    colframe=blue!50!black, colback=blue!5!white, coltitle=white,
    fonttitle=\bfseries, arc=1mm, boxsep=2pt, fontupper=\small,
]
You are an agent in a virtual science school environment, tasked to interact with various elements.

\vspace{4pt}
\textbf{Available commands:}
\begin{tcolorbox}[colback=white, colframe=gray!30, arc=0mm, boxsep=0pt, left=4pt, right=4pt, top=2pt, bottom=2pt]
\ttfamily\footnotesize
\textbf{Manipulation:} open/close \{OBJ\}, pick up \{OBJ\}, put down \{OBJ\}, move \{OBJ\} to \{OBJ\}, pour \{OBJ\} into \{OBJ\}, dunk \{OBJ\} into \{OBJ\}, mix \{OBJ\}.\par
\textbf{Inspection:} look around, look at \{OBJ\}, look in \{OBJ\}, read \{OBJ\}.\par
\textbf{Device Operations:} activate \{OBJ\}, deactivate \{OBJ\}, use \{OBJ\} [on \{OBJ\}].\par
\textbf{Movement:} go to \{LOC\}.\par
\textbf{Miscellaneous:} eat \{OBJ\}, flush \{OBJ\}, focus on \{OBJ\}, wait [DURATION].\par
\textbf{Information:} task, inventory.
\end{tcolorbox}
where \texttt{\{OBJ\}} is an object, \texttt{\{LOC\}} a location, and \texttt{[DURATION]} a specified time.

\vspace{4pt}
For each of your turns, you will be given the observation of the last turn. You should first think about what to do, and then output the action for this turn.

\vspace{4pt}
\textbf{Response format:}
\begin{tcolorbox}[colback=white, colframe=gray!30, arc=0mm, boxsep=0pt, left=4pt, right=4pt, top=2pt, bottom=2pt]
\ttfamily\footnotesize
Thought: \textless your thought\textgreater\par
Action: \textless your next action\textgreater
\end{tcolorbox}
\end{tcolorbox}
\caption{ReAct agent prompt for ScienceWorld, enumerating six command groups (Manipulation, Inspection, Device Operations, Movement, Miscellaneous, Information) and a per-turn \texttt{Thought}/\texttt{Action} response format.}
\label{fig:prompt_agent_react_sw}
\end{figure*}

\begin{figure*}[t]
\centering
\begin{tcolorbox}[
    title=\textbf{Prompt Template for ReflAct Agent (ScienceWorld)},
    colframe=blue!50!black, colback=blue!5!white, coltitle=white,
    fonttitle=\bfseries, arc=1mm, boxsep=2pt, fontupper=\small,
]
You are an agent in a virtual science school environment, tasked to interact with various elements. The available commands match the ReAct prompt above (Figure~\ref{fig:prompt_agent_react_sw}).

\vspace{4pt}
For each of your turns, you should first reflect in one sentence on the agent's state in relation to the task goal, and then output the action for this turn.

\vspace{4pt}
\textbf{Response format:}
\begin{tcolorbox}[colback=white, colframe=gray!30, arc=0mm, boxsep=0pt, left=4pt, right=4pt, top=2pt, bottom=2pt]
\ttfamily\footnotesize
Reflection: \textless your reflection\textgreater\par
Action: \textless your next action\textgreater
\end{tcolorbox}
\end{tcolorbox}
\caption{ReflAct agent prompt for ScienceWorld, inheriting the six ScienceWorld command groups.}
\label{fig:prompt_agent_reflact_sw}
\end{figure*}

\label{sec:appendix:prompts:wm}

\begin{figure*}[t]
\centering
\begin{tcolorbox}[
    title=\textbf{Prompt Template for Zero-Shot World Model},
    colframe=blue!50!black, colback=blue!5!white, coltitle=white,
    fonttitle=\bfseries, arc=1mm, boxsep=2pt, fontupper=\small,
]
You are a world model for the \{env\_name\} environment. Given the agent's current observation, its proposed action, and the task goal, predict what the agent will observe next.

\vspace{4pt}
\textbf{Output format:} a single paragraph starting with ``Prediction:'' describing the next observation in the style of the environment's own text output.

\vspace{4pt}
\textbf{User message:}
\begin{tcolorbox}[colback=white, colframe=gray!30, arc=0mm, boxsep=0pt, left=4pt, right=4pt, top=2pt, bottom=2pt]
\ttfamily\footnotesize
Task goal: \{task\_goal\}\par\medskip
Current observation: \{observation\}\par\medskip
Proposed action: \{action\}
\end{tcolorbox}
\end{tcolorbox}
\caption{Zero-Shot world model prompt.}
\label{fig:prompt_wm_zeroshot}
\end{figure*}

\begin{figure*}[t]
\centering
\begin{tcolorbox}[
    title=\textbf{Prompt Template for RAWM-$\phi$ World Model},
    colframe=blue!50!black, colback=blue!5!white, coltitle=white,
    fonttitle=\bfseries, arc=1mm, boxsep=2pt, fontupper=\small,
]
You are a world model. Given the agent's current state and proposed action, predict what the agent will observe next.

\vspace{4pt}
\textbf{Output format:} a single paragraph starting with ``Prediction:'' describing the next observation in the style of the environment's own text output.

\vspace{4pt}
\textbf{System-prompt block (retrieved transitions, prepended above the output directive):}
\begin{tcolorbox}[colback=white, colframe=gray!30, arc=0mm, boxsep=0pt, left=4pt, right=4pt, top=2pt, bottom=2pt]
\ttfamily\footnotesize
\#\# Retrieved similar past transitions\par
\textbf{Transition 1:}\par
~~State: \{state\_1\}\par
~~Action: \{action\_1\}\par
~~Observation: \{observation\_1\}\par
\textbf{Transition 2:} \ldots
\end{tcolorbox}
\end{tcolorbox}
\caption{RAWM-$\phi$ \cite{yang2025rawm} world model prompt, augmented with top-$k$ transitions retrieved by cosine similarity over \texttt{Qwen3-Embedding-8B} embeddings of the live $(s_t,a_t)$ query against the fixed Word2World corpus.}
\label{fig:prompt_wm_rawm}
\end{figure*}

\begin{figure*}[t]
\centering
\begin{tcolorbox}[
    title=\textbf{Prompt Template for ITP-I World Model},
    colframe=blue!50!black, colback=blue!5!white, coltitle=white,
    fonttitle=\bfseries, arc=1mm, boxsep=2pt, fontupper=\small,
]
You are a world model for the \{env\_name\} environment. Given an action/observation history, imagine the next few steps, describing likely observations and key objects.

\vspace{4pt}
\textbf{User message (one-step imagination contract for prediction; multi-step for planning):}
\begin{tcolorbox}[colback=white, colframe=gray!30, arc=0mm, boxsep=0pt, left=4pt, right=4pt, top=2pt, bottom=2pt]
\ttfamily\footnotesize
Goal: \{task\_goal\}\par\medskip
Current state: \{observation\}\par\medskip
Action: \{action\}
\end{tcolorbox}
Prediction-evaluation cells restrict ITP-I to a single imagined step so all methods share the next-observation target. Agent-planning cells let the world model select $k\in\{0,\ldots,5\}$ and return the corresponding $k$-step imagined future enclosed by \texttt{<Foresight>\ldots</Foresight>}.
\end{tcolorbox}
\caption{ITP-I world model prompt following the Imagine-then-Plan \cite{liu2026imaginethenplan} inference, with horizon $k$ fixed at $1$ for World Model Prediction evaluation and selected by the model during Agent Planning.}
\label{fig:prompt_wm_itp}
\end{figure*}

\begin{figure*}[t]
\centering
\begin{tcolorbox}[
    title=\textbf{Prompt Template for Episodic Memory World Model},
    colframe=blue!50!black, colback=blue!5!white, coltitle=white,
    fonttitle=\bfseries, arc=1mm, boxsep=2pt, fontupper=\small,
]
You are a world model for the \{env\_name\} environment. Given the agent's current observation, its proposed action, and the task goal, predict what the agent will observe next.

If a ``\#\# Retrieved similar past transitions'' section is provided, use those transitions as analogies for what can change after this action. Let the retrieved transitions guide what changes, but keep the prediction consistent with the current observation and task goal.

\vspace{4pt}
\textbf{Output format:} a single paragraph starting with ``Prediction:'' describing the next observation in the style of the environment's own text output.

\vspace{4pt}
\textbf{Grounding block (Episodic Memory):}
\begin{tcolorbox}[colback=white, colframe=gray!30, arc=0mm, boxsep=0pt, left=4pt, right=4pt, top=2pt, bottom=2pt]
\ttfamily\footnotesize
\#\# Retrieved similar past transitions\par
\textbf{Transition 1:} (State, Action, Observation)\par
\ldots\par
\textbf{Transition $k_{M_E}$:} (State, Action, Observation)
\end{tcolorbox}
\end{tcolorbox}
\caption{Episodic Memory world model prompt, augmented with top-$k_{M_E}$ action-keyed transitions retrieved from $M_E$ via Jaccard similarity over actions and prepended as a grounding block.}
\label{fig:prompt_wm_episodic}
\end{figure*}

\begin{figure*}[t]
\centering
\begin{tcolorbox}[
    title=\textbf{Prompt Template for Semantic Memory World Model},
    colframe=blue!50!black, colback=blue!5!white, coltitle=white,
    fonttitle=\bfseries, arc=1mm, boxsep=2pt, fontupper=\small,
]
You are a world model for the \{env\_name\} environment. Given the agent's current observation, its proposed action, and the task goal, predict what the agent will observe next.

If a ``\#\# Frame Axioms and Persistence Rules'' section appears above, treat each rule as a constraint: do not predict any change to a property the rules say does not change for the action being taken. Use the rules to filter out spurious changes.

\vspace{4pt}
\textbf{Output format:} a single paragraph starting with ``Prediction:'' describing the next observation in the style of the environment's own text output.

\vspace{4pt}
\textbf{Grounding block (Semantic Memory):}
\begin{tcolorbox}[colback=white, colframe=gray!30, arc=0mm, boxsep=0pt, left=4pt, right=4pt, top=2pt, bottom=2pt]
\ttfamily\footnotesize
\#\# Frame Axioms and Persistence Rules\par
\textbf{Rule 1:} \{rule\_text\} \; (evidence $e_1$)\par
\ldots\par
\textbf{Rule $|M_S|$:} \{rule\_text\} \; (evidence $e_{|M_S|}$)
\end{tcolorbox}
\end{tcolorbox}
\caption{Semantic Memory world model prompt, grounded by mismatch-derived persistence rules from $M_S$ rendered as frame axioms and ranked by accumulated evidence score.}
\label{fig:prompt_wm_semantic}
\end{figure*}

\begin{figure*}[t]
\centering
\begin{tcolorbox}[
    title=\textbf{Prompt Template for \textsc{WorldEvolver}},
    colframe=blue!50!black, colback=blue!5!white, coltitle=white,
    fonttitle=\bfseries, arc=1mm, boxsep=2pt, fontupper=\small,
]
You are a world model for the \{env\_name\} environment. Given the agent's current observation, its proposed action, and the task goal, predict what the agent will observe next.

If a ``\#\# Frame Axioms and Persistence Rules'' section appears above, treat each rule as a constraint: do not predict any change to a property the rules say does not change for the action being taken.

If a ``\#\# Retrieved similar past transitions'' section is provided, use those transitions as analogies for what can change after this action. Let the retrieved transitions guide what changes, and use the frame axioms / persistence rules to filter out spurious changes.

\vspace{4pt}
\textbf{Output format:} a single paragraph starting with ``Prediction:'' describing the next observation in the style of the environment's own text output.

\vspace{4pt}
\textbf{Grounding blocks (Semantic above, Episodic below):}
\begin{tcolorbox}[colback=white, colframe=gray!30, arc=0mm, boxsep=0pt, left=4pt, right=4pt, top=2pt, bottom=2pt]
\ttfamily\footnotesize
\#\# Frame Axioms and Persistence Rules\par
\{rules from $M_S$\}\par\medskip
\#\# Retrieved similar past transitions\par
\{$k_{M_E}$ transitions from $M_E$\}
\end{tcolorbox}
\end{tcolorbox}
\caption{\textsc{WorldEvolver} world model prompt combining Episodic and Semantic Memory grounding blocks, with \textit{foresight}-based confidence filtering applied post-generation without modifying the prompt.}
\label{fig:prompt_wm_worldevolver}
\end{figure*}

\begin{figure*}[t]
\centering
\begin{tcolorbox}[
    title=\textbf{Prompt Template for Observation Factorizer},
    colframe=blue!50!black, colback=blue!5!white, coltitle=white,
    fonttitle=\bfseries, arc=1mm, boxsep=2pt, fontupper=\small,
]
\textbf{System message:}
\begin{tcolorbox}[colback=white, colframe=gray!30, arc=0mm, boxsep=0pt, left=4pt, right=4pt, top=2pt, bottom=2pt]
\ttfamily\footnotesize
You extract compact world-state triples from text observations. Return only valid JSON; do not explain.
\end{tcolorbox}

\vspace{4pt}
\textbf{User message:}
\begin{tcolorbox}[colback=white, colframe=gray!30, arc=0mm, boxsep=0pt, left=4pt, right=4pt, top=2pt, bottom=2pt]
\ttfamily\footnotesize
Extract (subject, predicate, object) triples that describe the world state from the observation below. Output ONLY a JSON array of triples, each triple as a 3-element array of strings.\par\medskip
Environment: \{env\}\par\medskip
Observation: \{obs\}\par\medskip
Respond with JSON array only, no other text. Example output format:\par
[[``fridge 1'', ``is'', ``open''], [``mug 1'', ``on'', ``countertop 2''], [``water 1'', ``state'', ``liquid'']]
\end{tcolorbox}
\end{tcolorbox}
\caption{Observation factorizer prompt, converting predicted and gold observations into factorized triples whose set difference determines whether Semantic Memory identifies a mismatch.}
\label{fig:prompt_factorizer}
\end{figure*}

\begin{figure*}[t]
\centering
\begin{tcolorbox}[
    title=\textbf{Prompt Template for Preservation-Rule Extractor},
    colframe=blue!50!black, colback=blue!5!white, coltitle=white,
    fonttitle=\bfseries, arc=1mm, boxsep=2pt, fontupper=\small,
]
\textbf{System message:}
\begin{tcolorbox}[colback=white, colframe=gray!30, arc=0mm, boxsep=0pt, left=4pt, right=4pt, top=2pt, bottom=2pt]
\ttfamily\footnotesize
You are extracting PRESERVATION RULES for the \{env\} text environment from world model prediction mismatches.\par\medskip
A preservation rule says what should stay the same after an action. Look for cases where the prediction changed a fact, but the gold next observation shows that the fact did not change.\par\medskip
Return one JSON object per general rule:\par
- key: a short lowercase dot-separated name such as ``examine.object\_location''. Use the same key for the same rule across different mismatches.\par
- text: one clear sentence stating the rule generically for this environment, not for one specific object instance.\par\medskip
Example:\par
[\{``key'':``examine.object\_location'',``text'':``Examining an object does not move it.''\}]\par\medskip
Only output rules about facts that should NOT have changed. If the batch does not show a reusable preservation rule, output []. Do not invent rules to fill the list.\par\medskip
Output strict JSON only: a single list of objects with key and text fields. No markdown fences and no commentary.
\end{tcolorbox}

\vspace{4pt}
\textbf{User message (rendered over a mismatch batch of size $k_{M_S}$):}
\begin{tcolorbox}[colback=white, colframe=gray!30, arc=0mm, boxsep=0pt, left=4pt, right=4pt, top=2pt, bottom=2pt]
\ttfamily\footnotesize
Environment: \{env\}\par\medskip
From the following mismatches, extract preservation rules (what does NOT change after the action). Return strict JSON only: a JSON list of objects.\par\medskip
\textbf{Mismatch 1:}\par
~~State: \{state\_1\}\par
~~Action: \{action\_1\}\par
~~Prediction: \{prediction\_1\}\par
~~Gold next observation: \{gold\_next\_1\}\par\medskip
\textbf{Mismatch 2:} \ldots
\end{tcolorbox}
\end{tcolorbox}
\caption{Preservation-rule extractor prompt, processing a batch of $k_{M_S}$ mismatches per Semantic Memory update and returning a JSON rule list appended to $M_S$ for use in \textsc{WorldEvolver} grounding.}
\label{fig:prompt_rule_extractor}
\end{figure*}

\end{document}